\pdfoutput=1
\documentclass[11pt]{article}

\usepackage[preprint]{acl}

\usepackage{times}
\usepackage{latexsym}
\usepackage{float}
\usepackage[T1]{fontenc}
\usepackage[utf8]{inputenc}

\usepackage{microtype}

\usepackage{inconsolata}

\usepackage{graphicx}

\title{Reliably Bounding False Positives: A Zero-Shot Machine-Generated Text Detection Framework via Multiscaled Conformal Prediction}

\author{Xiaowei Zhu\textsuperscript{1,2}, Yubing Ren\textsuperscript{1,2}\thanks{Corresponding author.}, Yanan Cao\textsuperscript{1,2}, \\ \textbf{Xixun Lin\textsuperscript{1,2}, Fang Fang\textsuperscript{1,2}, Yangxi Li\textsuperscript{3}}\\
        \textsuperscript{1}Institute of Information Engineering, Chinese Academy of Sciences, Beijing, China \\ 
        \textsuperscript{2}School of Cyber Security, University of Chinese Academy of Sciences, Beijing, China \\
        \textsuperscript{3}National Computer Network Emergency Response Technical Team, Coordination Center of China, Beijing, China \\
        \texttt{\{zhuxiaowei\}@iie.ac.cn}}

\usepackage{amsmath}
\usepackage{amssymb}
\usepackage{booktabs} 
\usepackage{colortbl}
\usepackage{multirow}
\usepackage{float}
\usepackage{svg}
\usepackage{algorithm}
\usepackage{algorithmic}
\usepackage{subcaption}
\usepackage{graphicx}
\usepackage{amsthm}
\usepackage{array}
\usepackage{tabularx}
\usepackage{dblfloatfix} % 导言区加载此宏包
\usepackage{hyperref}
\usepackage{adjustbox}
\theoremstyle{plain} % 标题为粗体，正文为斜体（默认样式）
\newtheorem{theorem}{Theorem}
\newtheorem{corollary}{Corollary}
\newtheorem{observation}{Observation}
\newcommand{\cmark}{\textcolor{blue}{\textbf{\large\checkmark}}} % √ 绿色
\newcommand{\xmark}{\textcolor{red}{\textsf{X}}}    % X 红色

\begin{document}
\maketitle
\begin{abstract}
The rapid advancement of large language models has raised significant concerns regarding their potential misuse by malicious actors. As a result, developing effective detectors to mitigate these risks has become a critical priority. However, most existing detection methods focus excessively on detection accuracy, often neglecting the societal risks posed by high false positive rates (FPRs). This paper addresses this issue by leveraging Conformal Prediction (CP), which effectively constrains the upper bound of FPRs. While directly applying CP constrains FPRs, it also leads to a significant reduction in detection performance. To overcome this trade-off, this paper proposes a Zero-Shot Machine-Generated Text Detection Framework via Multiscaled Conformal Prediction (MCP), which both enforces the FPR constraint and improves detection performance. This paper also introduces RealDet, a high-quality dataset that spans a wide range of domains, ensuring realistic calibration and enabling superior detection performance when combined with MCP. Empirical evaluations demonstrate that MCP effectively constrains FPRs, significantly enhances detection performance, and increases robustness against adversarial attacks across multiple detectors and datasets.
\end{abstract}

\section{Introduction}

The rapid advancement of large language models (LLMs) has led to the generation of fluent, natural, and high-quality text that increasingly resembles human-written text. LLMs are being leveraged to enhance productivity across various domains, including news reporting, storytelling, and academic research \citep{Alshater2022ExploringTR, Yuan2022WordcraftSW, cnet2023ai}, significantly contributing to both industrial and academic progress. However,  this same capability has also opened the door for misuse, with malicious actors exploiting LLMs to generate fake news \cite{ahmed2021detectingfakenewsusing}, spam \cite{9121286}, malicious reviews \cite{Adelani2019GeneratingSF}, and other harmful contents that pose substantial risks to society. As a result, developing advanced Machine-Generated Text (MGT) detectors has become an urgent necessity.

\begin{figure}[t]
\centering
\counterwithout{figure}{section}
\resizebox{\linewidth}{!}{
\includegraphics{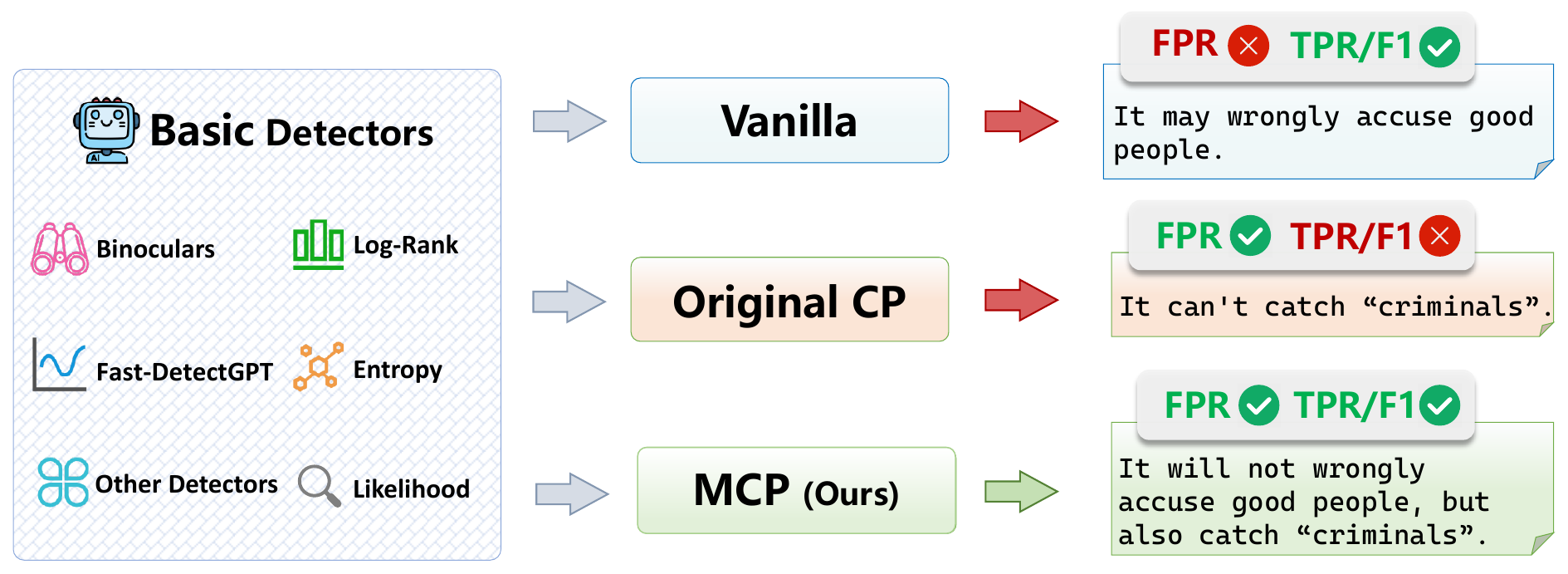}
}
\caption{Detection performance of detectors under different framework configurations.}
\label{fig1}
\end{figure}

Researchers have proposed numerous methods for MGT detection, including zero-shot detectors based on statistical metrics \citep{bao2024fastdetectgpt, hans2024spotting, pmlr-v202-mitchell23a} and supervised detectors fine-tuned on pretrained models \citep{solaiman2019release, Conneau2019UnsupervisedCR}. However, these approaches excessively emphasize detection accuracy while neglecting the potential societal harm caused by high false positive rates (FPRs). This concern is consistent with the findings of \citet{dugan-etal-2024-raid}, who highlighted that existing detectors often exhibit dangerously high FPRs under default thresholds. Detectors with high FPRs are impractical for real-world applications, as they fail to provide reliable guidance to users.

%In this paper, we aim to leverage conformal prediction (CP) \cite{Vovk1999MachineLearningAO} to address the challenges above. CP provides statistical guarantees with minimal assumptions, ensuring the prediction contains the ground truth. Specifically, it can offer an upper bound constraint on the FPR for MGT detection. As illustrated in Figure~\ref{fig1}, directly applying the CP framework effectively constrains the upper bound of the detector's FPR. While imposing this constraint improves the detection of misclassified human-written text (HWT), it may also allow certain machine-generated text to evade detection, significantly degrading the detector’s performance, which is counterproductive. Based on these, we propose a Zero-Shot Machine-Generated Text Detection Framework via \textbf{M}ultiscaled \textbf{C}onformal \textbf{P}rediction (\textbf{MCP})—a solution that not only effectively constrains the upper bound of the FPR but also enhances the detector’s original performance without any training.
In this paper, we propose leveraging conformal prediction (CP) \cite{Vovk1999MachineLearningAO} to address the challenges of high false positive rates (FPRs) in machine-generated text (MGT) detection. CP provides an upper bound on the FPR, ensuring that the detection results are reliable. While directly applying CP can constrain the FPR, it may also allow certain machine-generated texts to evade detection, which would degrade overall detection performance. To address this issue, we propose a novel solution, the Zero-Shot Machine-Generated Text Detection Framework via \textbf{M}ultiscaled \textbf{C}onformal \textbf{P}rediction (\textbf{MCP}). MCP not only effectively constrains the FPR but also improves detection performance without additional training.

%In our framework, we leverage the text distribution from the calibration set to generate detection results while adhering to a specified upper bound on the FPR. The theoretical foundation of CP guarantees the effectiveness of constraining the FPR. Our in-depth investigation into performance degradation reveals the need for an innovative multiscaled optimization, which effectively enhances detection performance. The calibration data within MCP should align with the realistic distribution of HWT, but most existing datasets fall short. We introduce \textbf{RealDet}, a high-quality benchmark to simulate \textbf{Real}-World Scenarios for MGT \textbf{Det}ection. 
The MCP framework operates in the following high-level manner: First, we sample both calibration and test sets from the target datasets. Next, we select a baseline detector and compute its nonconformity scores. From the calibration set’s nonconformity scores, we derive multiscaled quantiles, which act as thresholds for detection. These quantiles ensure that the FPR is constrained. Finally, we apply these thresholds to detect MGT instances in new, unseen data. Additionally, we introduce \textbf{RealDet}, a high-quality benchmark dataset designed to simulate realistic scenarios for MGT detection. RealDet is essential for ensuring that the calibration set reflects the true distribution of human-written text (HWT), addressing the gap in existing datasets.

Extensive experiments consistently demonstrate that the MCP framework effectively constrains the upper bound of the FPR while simultaneously improving detection performance. In adversarial scenarios, the MCP significantly enhances robustness. Our contributions are summarized as:
\begin{itemize}
\item We are the first to introduce CP into MGT detection and provide an in-depth exploration of potential optimization mechanisms.
\item We propose MCP, a zero-shot detection framework that not only constrains the FPR upper bound but also improves detection performance and enhances robustness against adversarial attacks.
\item We construct RealDet, the large-scale and comprehensive bilingual benchmark,  consisting of 847k raw texts spanning 15 representative domains, 22 popular and powerful LLMs, and covering two adversarial attacks. %Utilizing RealDet as a calibration set within MCP demonstrates superior detection performance in real-world detection scenarios.
\end{itemize}

\section{Preliminary}\label{preliminary}

\paragraph{Conformal Prediction.} 
Conformal prediction \citep{Vovk2005AlgorithmicLI, Papadopoulos2002InductiveCM, Lei2014DistributionfreePB} is a statistical learning framework that generates reliable prediction without training. It provides statistical guarantees for the coverage of the ground truth assuming only data exchangeability. The workflow is:

1. Split the data into a calibration set $D_{cal}$ and a test set $D_{test}$, with $D_{cal}$ containing 
$n$ instances.

2. Given a model taking input $x$ and producing output $y$. Then define a nonconformity score \( s(x, y) \in \mathbb{R} \), where larger scores encode worse agreement between $x$ and $y$.

3. Compute quantile \(\hat{q}\) of $s$ derived from $D_{cal}$:
        \begin{equation}
        \scalebox{0.95}{$\hat{q}=\operatorname{quantile}\left(s_1, \ldots, s_n; \frac{\lceil(n+1)(1-\alpha)\rceil}{n}\right)$.}
        \label{eq:2}
        \end{equation}

4. Using $\hat{q}$ as the prediction threshold to predict each test instance:
    \begin{equation}
    \mathcal{C}\left(X_{\text {test }}\right)=\left\{y: s\left(X_{\text {test }}, y\right) \leq \hat{q}\right\} .
    \end{equation}

% \begin{itemize}
%     \item[1.] Split the data into a calibration set $D_{cal}$ and a test set $D_{test}$. 
%     \item[2.] Given a model taking input $x$ and producing output $y$. Then define a nonconformity score \( s(x, y) \in \mathbb{R} \), where larger scores encode worse agreement between $x$ and $y$.
%     \item[3.] Compute quantile \(\hat{q}\) of $s$ derived from $D_{cal}$:
%         \begin{equation}
%         \scalebox{0.95}{$\hat{q}=\operatorname{quantile}\left(s_1, \ldots, s_n; \frac{\lceil(n+1)(1-\alpha)\rceil}{n}\right)$.}
%         \label{eq:2}
%         \end{equation}
%     \item[4.] Using $\hat{q}$ as the prediction threshold to predict each test instance:
%     \begin{equation}
%     \mathcal{C}\left(X_{\text {test }}\right)=\left\{y: s\left(X_{\text {test }}, y\right) \leq \hat{q}\right\} .
%     \end{equation}
% \end{itemize}
\begin{theorem}
\textbf{Conformal coverage guarantee} \cite{Vovk1999MachineLearningAO}. Suppose the calibration set \((X_i, Y_i)_{i=1, \dots, n}\) and the new instance \((X_{\text{test}}, Y_{\text{test}})\) are independent and identically distributed (i.i.d.). Then, the following holds:
\begin{equation}
P\left(Y_{\text {test }} \in \mathcal{C}\left(X_{\text {test }}\right)\right) \geq 1-\alpha .
\vspace{-5mm}
\end{equation}
\label{theorem 1}
\end{theorem}

\paragraph{MGT Detection within CP.} 
Given $n$ human-written texts $(X_1, X_2, \ldots, X_n)$ as a calibration set, we are tasked to predict a new instance $X_{test}$ is human-written or machine-generated. Based on the output of detector $Det$, we define a nonconformity score $s \in [0, 1]$, where a larger score indicates a lower probability that the text is human-written. Then we compute quantile \(\hat{q}\) according to Equation \ref{eq:2} and use \(\hat{q}\) as the threshold to make prediction:
\begin{equation}
\scalebox{0.90}{$
\mathcal{C}(X_{test})= \begin{cases}\text { HWT, } & s(X_{test}) \leq \hat{q} \\ \text { MGT, } & s(X_{test})>\hat{q} .\end{cases}$}
\end{equation}

This gives the guarantee in Theorem~\ref{theorem 1} that no more than $\alpha$ fraction of future human-written texts will be misclassified as machine-generated, i.e. FPR$\leq\alpha$.

\begin{figure*}[t]
\centering
\counterwithout{figure}{section}
\resizebox{\linewidth}{!}{
\includegraphics{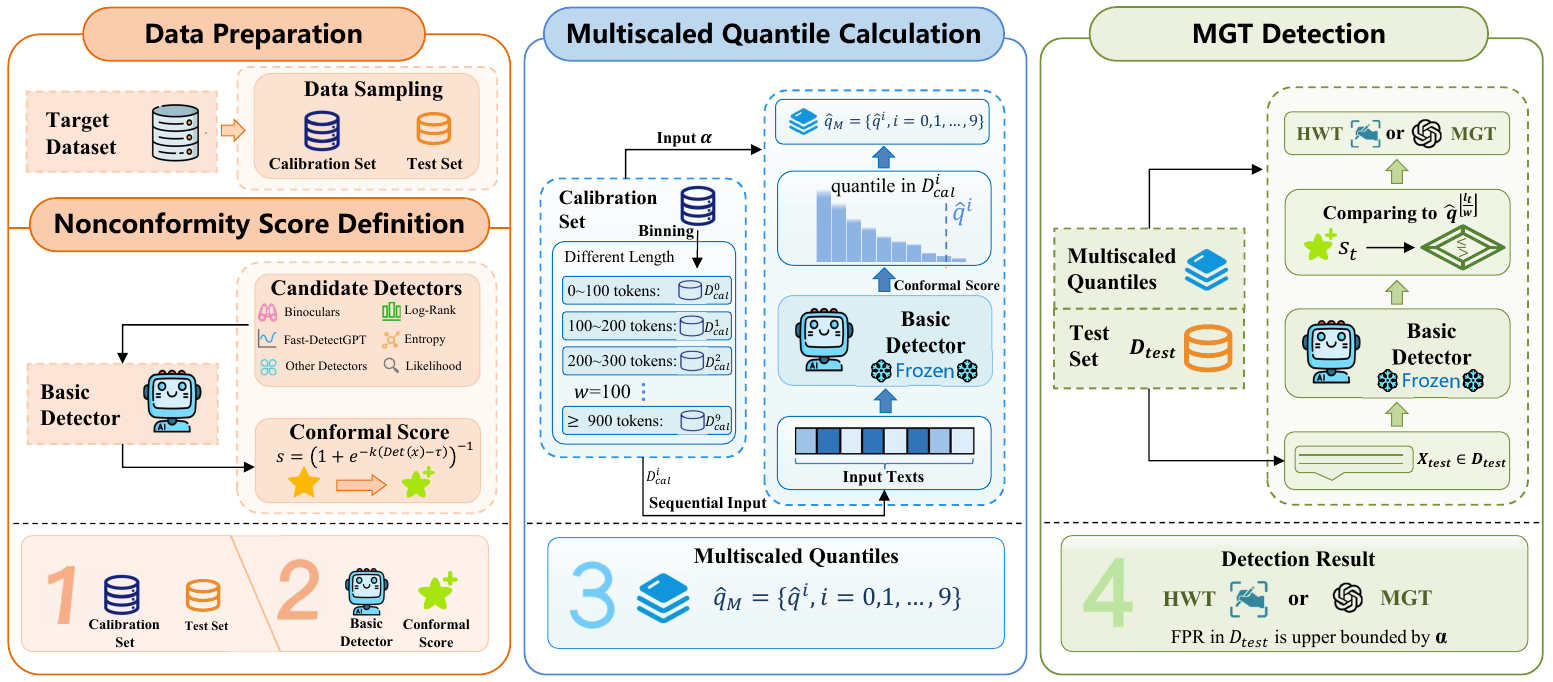}
}
\caption{\textbf{The MCP Framework.} The prediction process consists of four parts, which are executed sequentially: data preparation, nonconformity score definition, multiscaled quantiles calculation, and MGT Detection.}
\label{fig2}
\vspace{-1mm}
\end{figure*}

\section{Multiscaled Conformal Prediction}

% MCP differentiates itself from other threshold calibration methods by its core advantage: improving performance while ensuring control over the FPR.

% Our framework, MCP, can effectively constrain the upper bound of the FPR in MGT detection while improving detection performance without requiring additional training. 
Figure~\ref{fig2} illustrates the MCP prediction process. First, we sample calibration and test sets from the target datasets. Next, we determine a basic detector and define its nonconformity scores. Subsequently, we derive multiscaled quantiles from the calibration set’s nonconformity scores. Finally, we apply the multiscaled quantiles as the threshold to perform MGT detection on new instances.

\subsection{Data Preparation}
We sampled from the target dataset to create calibration and test sets, where the calibration set $D_{cal}$ consists entirely of human-written texts, while the test set $D_{test}$ includes both human-written and machine-generated texts. Sampling from the same dataset ensures that the human-written text in both the calibration and test sets is independent and identically distributed (i.i.d.).

% Additionally, we perform length-aware binning on the calibration set, dividing it into multiple subsets $\{S_C^1, S_C^2, \ldots, S_C^K\}$, each representing a collection of texts within a specific length interval. The binning strategy we adopt determines the number of subsets $K$. In our framework, we typically use an equal-width binning strategy, where the maximum input text length $L_\phi$ is divided into corresponding length intervals of fixed width $w$ as follows:
% \begin{equation}
% K = \lfloor {\frac{L_\phi}{w}} \rfloor .
% \end{equation}

\subsection{Nonconformity Score Definition}
First, we determine the basic detector $Det$, whose selection is highly flexible and can include most detectors designed for MGT detection. Then we define a nonconformity score function $s(\cdot)$ that converts the output of the basic detector $Det(x)$ into a nonconformity score $s$ for making predictions:
\begin{equation}
s = (1 + e^{-k(Det(x) - \tau)})^{-1},
\end{equation}
where $\tau$ represents the default threshold of the basic detector, and $k$ takes a value of either -1 or 1. A larger value of $s$ signifies a lower probability that the input text is human-written text.
% , depending on whether the basic detector's output for human-written text is expected to be greater or smaller.

% For example, in Binoculars, $\tau_\phi = 0.901$ and $k = -1$.

\begin{table*}[!b]
\resizebox{\linewidth}{!}{%
    \centering
    \setlength{\tabcolsep}{8pt} % 调整列之间的间距
    \renewcommand{\arraystretch}{1} % 调整行高，增加表格的空白，使其更易读
    \begin{tabular}{
        l| % 左对齐
        >{\centering\arraybackslash}m{2cm}|
        >{\centering\arraybackslash}m{2cm}| % 居中对齐，固定宽度
        >{\centering\arraybackslash}m{2cm}| 
        >{\centering\arraybackslash}m{2cm}| 
        >{\centering\arraybackslash}m{2cm}}
        \toprule
        % \rowcolor{gray!25} % 浅灰色表头
        \textbf{Dataset} & \textbf{Origin Size} & \textbf{Domain Coverage} & \textbf{Model Coverage} & \textbf{Multilingual Coverage} & \textbf{Adversarial Coverage} \\
        \midrule
        % \rowcolor{gray!10} % 淡灰色的偶数行
        TuringBench \cite{uchendu-etal-2021-turingbench-benchmark} & 200k & \xmark & \cmark(10) & \xmark & \xmark \\
        HC3 \cite{guo2023closechatgpthumanexperts} & 26.9k & \cmark(5) & \xmark & \cmark & \xmark \\
        CHEAT \cite{DBLP:journals/corr/abs-2304-12008} & 50k & \xmark & \xmark & \xmark & \cmark \\
        % \rowcolor{gray!15} % 略深的灰色
        MGTBench \cite{he2024mgtbenchbenchmarkingmachinegeneratedtext} & 18.5k & \cmark(3) & \cmark(5) & \xmark & \xmark \\
        M4 \cite{wang-etal-2024-m4} & 122k & \cmark(5) & \cmark(7) & \cmark & \xmark \\
        % \rowcolor{gray!10} % 淡灰色的偶数行
        
        MAGE \cite{li-etal-2024-mage} & 447k & \cmark(10) & \cmark(9) & \xmark & \xmark \\
        % \rowcolor{gray!15} % 略深的灰色
        RAID \cite{dugan-etal-2024-raid} & 570k & \cmark(8) & \cmark(8) & \xmark & \cmark \\
        \midrule
        % \rowcolor{cyan!10} % 清新的浅蓝色
        \textbf{RealDet (Ours)} & \textbf{847k} & \cmark\textbf{(15)} & \cmark\textbf{(22)}& \cmark & \cmark \\
        \bottomrule
    \end{tabular}}
    \caption{Comparison of open-source datasets in MGT detection. The “Origin size” refers to the number of \textbf{raw texts} without adversarial attacks. The “Model Coverage” column represents the count of \textbf{base models}.}
    % RealDet is the only dataset that spans multiple domains, models, and languages, while also incorporating adversarial attacks and covering a more diverse range of domains.
    \label{tab:dataset_comparison}
    \vspace{-4pt}
\end{table*}

\subsection{Multiscaled Quantile Calculation}

\paragraph{Problem in traditional quantile calculation within CP.}
%As discussed in Section~\ref{preliminary}, CP ensures that the FPR remains below a specified threshold $\alpha$. Based on the given definition, the traditional quantile can be calculated as shown in Equation~\ref{eq:2}. 
As illustrated in Figure~\ref{fig3}, while traditional computational approaches effectively control the FPR of prediction results, they do so at the significant cost of detection performance. This trade-off prevents the detection of the majority of machine-generated texts. Based on our data analysis, we observe the following:

\begin{figure}[t]
\centering
\counterwithout{figure}{section}
\resizebox{\linewidth}{!}{
\includegraphics{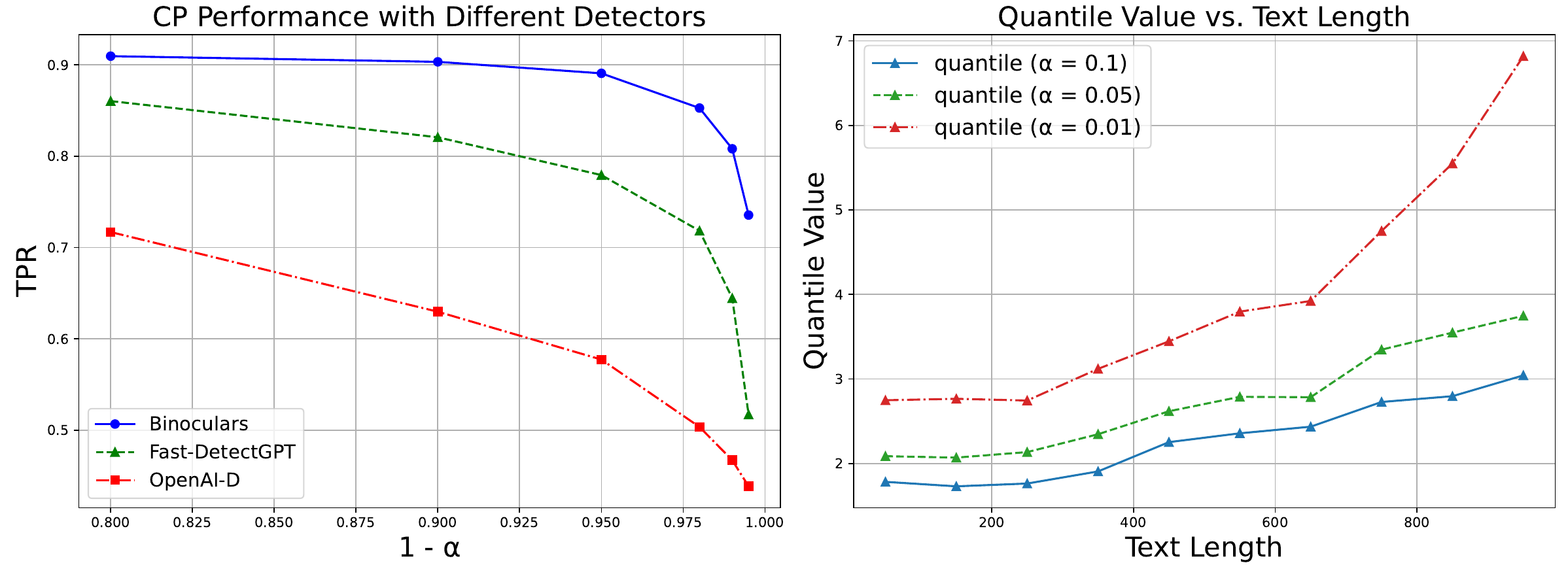}
}
\caption{Left: True Positive Rate (TPR) of different detectors with the CP as a function of $\alpha$.
Right: Quantile values calculated for different text length intervals.}
\label{fig3}
\vspace{-10pt}
\end{figure}

\begin{observation}
\normalfont
\textbf{Longer texts tend to have higher nonconformity scores.} Figure~\ref{fig3} shows that text length significantly influences the magnitude of the quantiles. Consequently, we further calculated the Pearson correlation coefficient $\rho_{l,s}$ between text length and the nonconformity score, finding that $\rho_{l,s}$ is close to 1, which indicates a strong positive correlation. Machine-generated texts with shorter lengths and lower nonconformity scores may remain undetected, leading to a significant decline in detection performance. 
\label{observation 1}
\end{observation}
% The formula for the Pearson correlation coefficient is given as follows:
% \begin{equation}
% \rho_{l,s} = \frac{\sum_{i=1}^{N} (l_i - \bar{l})(s_i - \bar{s})}{\sqrt{\sum_{i=1}^{N} (l_i - \bar{l})^2} \sqrt{\sum_{i=1}^{N} (s_i - \bar{s})^2}}.
% \label{eq:6}
% \end{equation}

\paragraph{Multiscaled quantiles calculation within MCP.} We incorporate the positive correlation between text length and nonconformity score into the prediction process. So we perform length-aware binning on the calibration set $D_{cal}$, dividing it into multiple subsets $\{D_{cal}^1, D_{cal}^2, \ldots, D_{cal}^K\}$, corresponding to a specific text length interval. We employ an equal-width binning strategy, partitioning the maximum input text length $L_{\text{max}}$ into fixed-width intervals of width $w$, as follows:
\begin{equation}
K = \lfloor {\frac{L_{\text{max}}}{w}} \rfloor .
\end{equation}

The multiscaled quantiles $\hat{q}_{M}$ are derived from nonconformity scores calculated over the subsets, each corresponding to different length intervals:
\begin{align} \label{Eq: quantile}
    \hat{q}_{M} &= \{ \hat{q}^i \mid \hat{q}^i = \operatorname{quantile}\left(s^i_1, s^i_2, \ldots, s^i_{n_i}; \right. \\
    & \nonumber \quad \left. \lceil({n_i}+1)(1-\alpha)\rceil{n_i}^{-1}\right),  i = 1, 2, \dots, K\},
\end{align}
where $n_i$ denotes the number of texts in $D_{cal}^i$, $s^i$ represents the nonconformity scores calculated from $D_{cal}^i$, and $\alpha$ denotes the desired upper bound of the FPR. By utilizing $\hat{q}_{M}$, we can select more appropriate quantiles for calibration across varying text lengths.

\subsection{MGT Detection}
For a new instance $X_{test}$ from the test set $D_{test}$, we classify it based on its nonconformity score $s_t$ and text length $l_t$. The detection result within MCP can be expressed as follows:
\begin{equation}
s_t = (1 + e^{-k(Det(X_\text{test}) - \tau)})^{-1},
\label{eq:8}
\end{equation}
\begin{equation}
\mathcal{C}(X_{test}) = \mathbb{I}(s_t > \hat{q}^{\lfloor {\frac{l_t}{w}} \rfloor}),
\end{equation}
where $\hat{q}^{\lfloor {\frac{l_t}{w}} \rfloor}$ represents the quantile within the corresponding length interval, and $\mathcal{C}(X_{test})$ denotes the detection result. $\mathcal{C}(X_{test})=0$ means that $X_{test}$ is human-written text, while $\mathcal{C}(X_{test})=1$ indicates that $X_{test}$ is machine-generated text. A detailed case study is in Appendix \ref{appendix: case study}.

\begin{corollary}
\label{corollary 1}
 The upper bound of the FPR for MGT detection within the MCP framework is $\alpha$. The detailed proof is provided in Appendix~\ref{appendix: proof}.
\end{corollary}

%While obtaining both the calibration and test sets from the same dataset is often feasible, such conditions may not hold in real-world detection scenarios. In the "wild" scenario, it may be challenging to collect calibration data from unseen text sources. Hence, constructing a calibration set that adequately represents the overall distribution of human-written texts is of critical importance within MCP.

\begin{figure*}[!t]
\centering
\counterwithout{figure}{section}
\resizebox{\linewidth}{!}{
\includegraphics{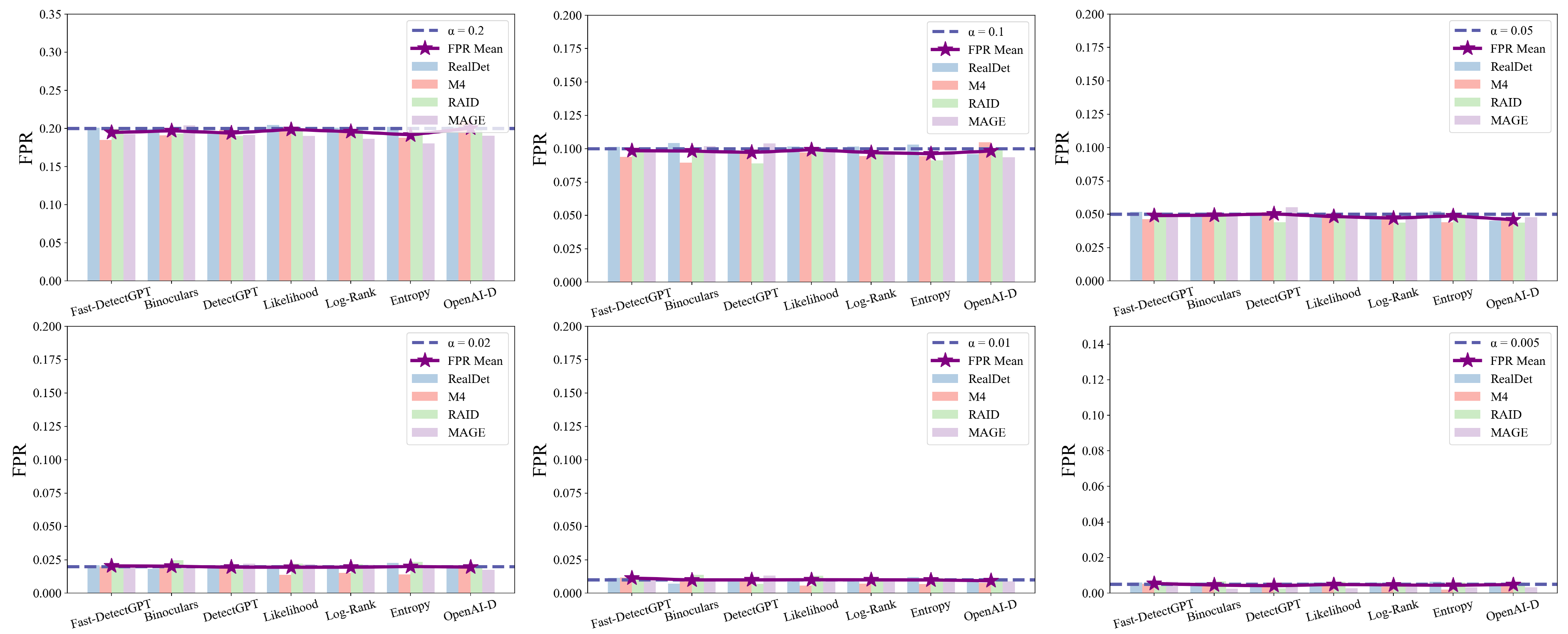}
}
\caption{The FPR of various detectors within the MCP framework across all datasets, after applying alpha constraints with values of alpha set to 0.2, 0.1, 0.05, 0.02, 0.01, and 0.005.}
\label{fig4}
\vspace{-10pt}
\end{figure*}

\begin{algorithm}[t]
\footnotesize
\caption{MCP Framework}
\begin{algorithmic}[1]
    \STATE \textbf{Input:} calibration set $D_{cal}$, test set $D_{test}$, basic detector $Det$, nonconformity score function $s(\cdot)$

    \STATE \textbf{Compute multiscaled quantiles} $\hat{q}_{M}$:

    \STATE Number of subsets $K \leftarrow \lfloor {\frac{L_{\text{max}}}{w}} \rfloor$
    
    \STATE Bin calibration set $\{D_{cal}^1,\ldots, D_{cal}^K\} \leftarrow D_{cal}$
    
    \FOR{$i = 1$ to $K$}
    
    \STATE $S^i \leftarrow \{s^i_1,\ldots, s^i_{n_i}\}$

    \STATE $\delta^i \leftarrow \lceil({n_i}+1)(1-\alpha)\rceil{n_i}^{-1}$
    
    \STATE  $\hat{q}^i \leftarrow \operatorname{quantile}(S^i; \delta^i)$
    
    \ENDFOR
    
    \STATE $\hat{q}_{M} \leftarrow \{ \hat{q}^i \mid  i = 1, 2, \dots, K\}$
    
    % \STATE \textbf{Output:} "0" if $X_\text{test}$ is a HWT, "1" otherwise
    
    \STATE \textbf{Detect:} $X_\text{test} \in D_{test}$

    \STATE $s_t \leftarrow s(X_\text{test})$
    
    \STATE $\mathcal{C}(X_{test}) \leftarrow \mathbb{I}(s_t > \hat{q}^{\lfloor {\frac{l_t}{w}} \rfloor})$

    \IF{$\mathcal{C}(X_{test})=0$}
        \STATE $X_\text{test}$ is a human-written text.
    \ELSE
        \STATE $X_\text{test}$ is a machine-generated text.
    \ENDIF
    
    % \RETURN $\mathcal{C}(X_{test})$
\end{algorithmic}
\end{algorithm}

\section{RealDet Dataset}
Existing datasets are limited in scope and exhibit domain-specific biases \citep{wu2024surveyllmgeneratedtextdetection, wu2024detectrl}, rendering them inadequate for representing human-written texts across all domains. We introduce the RealDet dataset and compare it with publicly available datasets in Table \ref{tab:dataset_comparison}. RealDet offers three key advantages: \textbf{(1) Comprehensive Domain Coverage.} RealDet spans \textbf{15} distinct textual domains, far exceeding existing datasets in domain diversity. \textbf{(2) Extensive Model Coverage.} RealDet is constructed using \textbf{22} popular and powerful LLMs, with the broadest range of base models currently. \textbf{(3) Large-scale Text Corpus.} RealDet includes over \textbf{847k} raw texts (excluding adversarial texts), with more than \textbf{113k} human-written texts, significantly surpassing other datasets in the size of raw texts. Furthermore, RealDet includes bilingual texts in both Chinese and English and adversarial texts involving paraphrasing and editing attacks.

\paragraph{Data Sources.}
To emulate the texts encountered in real-world detection scenarios, we carefully filtered the existing datasets and selected 15 representative data sources covering 6 writing tasks. (1) \textbf{Question Answering}: ELI5 \cite{fan-etal-2019-eli5}, WiKiQA \cite{yang-etal-2015-wikiqa}, Medical Dialog \cite{he2020meddialoglargescalemedicaldialogue}, FiQA \cite{10.1145/3184558.3192301}; (2) \textbf{News Article Writing}: XSum \cite{narayan-etal-2018-dont}, TLDR\footnote{https://huggingface.co/datasets/JulesBelveze/TLDR\_news}, BBC News \cite{Greene2006PracticalST}; (3) \textbf{Story Generation}: WritingPrompt \cite{fan-etal-2018-hierarchical}, ROC Stories \cite{mostafazadeh-etal-2016-corpus}; (4) \textbf{Review Expression}: Yelp \cite{NIPS2015_250cf8b5}, IMDB \cite{maas-etal-2011-learning}, CMV \cite{10.1145/2872427.2883081}; (5) \textbf{Academic Writing}: Abstracts\footnote{https://www.kaggle.com/datasets/spsayakpaul/arxiv-paper-abstracts}; (6) \textbf{Knowledge Explanation}: Wikipedia \cite{aaditya_bhat_2023}, SQuAD \cite{rajpurkar-etal-2016-squad}. More details are in Appendix~\ref{appendix: data sources}.

\paragraph{Model Set.}
We consider black-box models (service provider offers API access) and white-box models (open-source models are deployed locally), 22 in total. (1) \textbf{Black-box}: Deepseek-R1, GPT-4, ChatGPT, PaLM 2, Ernie Bot 3.5 turbo, Spark Desk 2.0, Qwen turbo, 360GPT S2 V9, Minimax abab 5.5; (2) \textbf{White-box}: LLaMA2-13B, ChatGLM2-6B, MOSS-moon-003, MPT-7B, InternLM-7B, Alpaca-7B, Guanaco-7B, Vicuna-13B, BLOOMz-7B, Falcon-7B, OPT-6.7B, Baichuan-13B, Flan-T5-XXL. Details are in Appendix~\ref{appendix: model supplement}.

\paragraph{Prompt Design.}
To collect machine-generated text for
each instance, we design three types of prompts to feed the LLMs. (1) \textbf{Continuation Writing}: ask LLMs to continue generation based on the first sentence of the original human-written text; (2) \textbf{Topical Writing}: ask LLMs to generate topic-specific texts (e.g., news article, paper abstract, etc.); (3) \textbf{Question-Answering}: ask LLMs to generate an answer based on a given question.  Specific prompts are in Appendix~\ref{appendix: prompt design}.

% \paragraph{Additional Details.}
% Furthermore, the RealDet dataset is a cross-linguistic dataset, encompassing both Chinese and English text data. The dataset also includes various types of adversarial attacks, such as text-editing attacks and paraphrasing attacks.

\begin{table*}[!t]
    \centering
    \resizebox{\textwidth}{!}{
    \large
    \begin{tabular}{llcccccccccccc}
        \toprule
        \textbf{\textit{Detector}} & \textbf{\textit{Algorithm}} & \textbf{$TP@20\%$} & \textbf{$F_1@20\%$} & \textbf{$TP@10\%$} & \textbf{$F_1@10\%$} & \textbf{$TP@5\%$} & \textbf{$F_1@5\%$} & \textbf{$TP@2\%$} & \textbf{$F_1@2\%$} & \textbf{$TP@1\%$} & \textbf{$F_1@1\%$} & \textbf{$TP@0.5\%$} & \textbf{$F_1@0.5\%$} \\ 
        \midrule
        \rowcolor[gray]{0.9} \multicolumn{14}{c}{\textbf{M4}} \\
        \multirow[t]{3}{*}{Fast-DetectGPT} & vanilla & 78.56 & 79.13 & 74.44 & 80.72 & 70.04 & 80.03 & 65.32 & 78.08 & 60.56 & 74.97 & 54.84 & 70.60 \\
                                  & \textbf{MCP}  & \textbf{79.56} & \textbf{80.35} & \textbf{75.24} & \textbf{81.50} & \textbf{71.72} & \textbf{81.33} & \textbf{67.36} & \textbf{79.55} & \textbf{62.44} & \textbf{76.33} & \textbf{58.64} & \textbf{73.67} \\
                                  \midrule
        \multirow[t]{3}{*}{Binoculars} & vanilla & 83.20 & 81.87 & 79.44 & 83.87 & 74.80 & 83.19 & 69.72 & 81.21 & 64.52 & 77.96 & 55.36 & 71.03 \\
                                  & \textbf{MCP} & \textbf{83.24} & \textbf{82.40} & \textbf{80.00} & \textbf{84.41} & \textbf{75.72} & \textbf{83.87} & \textbf{71.20} & \textbf{82.27} & \textbf{66.68} & \textbf{79.59} & \textbf{62.72} & \textbf{76.88} \\\midrule
        \rowcolor[gray]{0.9} \multicolumn{14}{c}{\textbf{RAID}} \\
        \multirow[t]{3}{*}{Fast-DetectGPT} & vanilla & 77.97 & 78.74 & 75.23 & 81.22 & 71.67 & 81.13 & 66.70 & 79.08 & 64.70 & 78.09 & 63.10 & 77.14 \\
                                  & \textbf{MCP} & \textbf{78.13} & \textbf{78.88} & \textbf{76.03} & \textbf{81.77} & \textbf{73.43} & \textbf{82.40} & \textbf{68.27} & \textbf{80.20} & \textbf{64.73} & \textbf{78.09} & \textbf{63.60} & \textbf{77.56} \\\midrule
        \multirow[t]{3}{*}{Binoculars} & vanilla & 78.40 & 79.03 & 76.47 & 82.02 & 74.53 & 83.03 & 70.17 & 81.50 & 67.50 & 80.12 & 64.17 & 77.94 \\
                                  & \textbf{MCP} & \textbf{78.50} & \textbf{79.19} & \textbf{76.57} & \textbf{82.21} & \textbf{74.83} & \textbf{83.19} & \textbf{72.37} & \textbf{82.78} & \textbf{70.33} & \textbf{81.93} & \textbf{66.07} & \textbf{79.25} \\\midrule
        \rowcolor[gray]{0.9} \multicolumn{14}{c}{\textbf{MAGE}} \\
        \multirow[t]{3}{*}{Fast-DetectGPT} & vanilla & 80.36 & 80.15 & 77.08 & 82.40 & 72.92 & 81.99 & 66.04 & 78.60 & 57.00 & 72.15 & 43.08 & 60.00 \\
                                  & \textbf{MCP} & \textbf{82.12} & \textbf{81.57} & \textbf{79.72} & \textbf{84.08} & \textbf{77.28} & \textbf{84.83} & \textbf{72.44} & \textbf{83.02} & \textbf{67.92} & \textbf{80.40} & \textbf{61.24} & \textbf{75.64} \\\midrule
        \multirow[t]{3}{*}{Binoculars} & vanilla & 85.12 & 83.01 & 84.00 & 86.62 & 82.56 & 88.04 & 74.60 & 84.50 & 56.04 & 71.37 & 28.52 & 44.20 \\
                                  & \textbf{MCP} & \textbf{85.12} & \textbf{83.63} & \textbf{84.04} & \textbf{86.74} & \textbf{82.68} & \textbf{88.07} & \textbf{77.36} & \textbf{86.32} & \textbf{75.80} & \textbf{85.77} & \textbf{73.32} & \textbf{84.49} \\\midrule
        \rowcolor[gray]{0.9} \multicolumn{14}{c}{\textbf{RealDet}} \\
        \multirow[t]{3}{*}{Likelihood} & vanilla & 83.60 & 82.11 & 79.70 & 84.03 & 76.08 & 84.03 & 67.38 & 79.56 & 58.98 & 73.73 & 36.92 & 53.73 \\
                                  & \textbf{MCP} & \textbf{84.10} & \textbf{82.23} & \textbf{80.64} & \textbf{84.53} & \textbf{76.66} & \textbf{84.50} & \textbf{70.06} & \textbf{81.43} & \textbf{62.24} & \textbf{76.21} & \textbf{52.60} & \textbf{68.67} \\\midrule
        \multirow[t]{3}{*}{Log-Rank} & vanilla & 84.74 & 82.80 & 81.30 & 84.99 & 77.96 & 85.22 & 70.48 & 81.73 & 61.50 & 75.69 & 45.74 & 62.55 \\
                                  & \textbf{MCP} & \textbf{85.12} & \textbf{82.95} & \textbf{81.88} & \textbf{85.27} & \textbf{78.36} & \textbf{85.63} & \textbf{72.22} & \textbf{82.93} & \textbf{65.84} & \textbf{78.90} & \textbf{58.28} & \textbf{73.36} \\\midrule
        \multirow[t]{3}{*}{Entropy} & vanilla & 68.28 & 72.51 & 46.44 & 59.37 & 30.24 & 44.72 & 14.26 & 24.53 & 6.58 & 12.23 & 3.18 & 6.13 \\
                                  & \textbf{MCP} & \textbf{70.60} & \textbf{74.01} & \textbf{48.16} & \textbf{60.79} & \textbf{31.92} & \textbf{46.55} & \textbf{17.68} & \textbf{29.48} & \textbf{9.38} & \textbf{16.97} & \textbf{5.66} & \textbf{10.65} \\\midrule
        \multirow[t]{3}{*}{DetectGPT} & vanilla & 71.10 & 74.42 & 55.14 & 66.78 & 38.20 & 53.35 & 19.88 & 32.62 & 10.42 & 18.70 & 5.96 & 11.19 \\
                                  & \textbf{MCP} & \textbf{73.18} & \textbf{75.92} & \textbf{57.82} & \textbf{69.18} & \textbf{39.88} & \textbf{55.17} & \textbf{21.94} & \textbf{35.44} & \textbf{13.98} & \textbf{24.33} & \textbf{6.78} & \textbf{12.64} \\\midrule
        \multirow[t]{3}{*}{OpenAI-D} & vanilla & 71.06 & 74.39  & 62.94 & 72.80 & 57.98 & 71.16 & 51.50 & 67.11 & 47.32 & 63.81 & 43.62 & 60.53\\
                                  & \textbf{MCP} & \textbf{79.40} & \textbf{79.54} & \textbf{67.74} & \textbf{76.40} & \textbf{59.14} & \textbf{72.28} & \textbf{53.98} & \textbf{69.18} & \textbf{49.70} & \textbf{65.94} & \textbf{45.50} & \textbf{62.34} \\\midrule
        \multirow[t]{3}{*}{Fast-DetectGPT} & vanilla & 86.02 & 83.52 & 81.86 & 85.34 & 77.46 & 84.91 & 72.00 & 82.76 & 63.74 & 77.38 & 51.22 & 67.52 \\
                                  & \textbf{MCP} & \textbf{87.10} & \textbf{84.07} & \textbf{84.24} & \textbf{86.68} & \textbf{80.86} & \textbf{86.93} & \textbf{76.86} & \textbf{85.90} & \textbf{73.20} & \textbf{83.97} & \textbf{69.32} & \textbf{81.59} \\\midrule
        \multirow[t]{3}{*}{Binoculars} & vanilla & 90.96 & 86.20 & 90.36 & 90.19 & 89.16 & 91.83 & 84.98 & 90.90 & 78.98 & 87.77 & 70.16 & 82.22 \\
                                  & \textbf{MCP} & \textbf{91.06} & \textbf{86.30} & \textbf{90.36} & \textbf{90.36} & \textbf{89.26} & \textbf{92.13} & \textbf{87.50} & \textbf{92.44} & \textbf{86.28} & \textbf{92.28} & \textbf{84.34} & \textbf{91.29} \\
        \bottomrule
    \end{tabular}
    }
    \caption{Main Experimental Results Across Various Detectors and Datasets. “Vanilla” refers to the detector's original configuration, whereas “MCP” denotes detectors with MCP framework.}
    \label{tab:main_experiment}
    \vspace{-10pt}
\end{table*}

\section{Experiments}

We conduct comprehensive experiments to thoroughly evaluate MCP, focusing on its ability to constrain the FPR, evaluate detection performance, test robustness against real-world attacks, investigate the impact of calibration data, and compare with other calibration methods.

\subsection{Experimental Setup}

\paragraph{Datasets.}
We evaluate MCP on RealDet and three representative datasets—M4 \cite{wang-etal-2024-m4}, RAID \cite{dugan-etal-2024-raid}, and MAGE \cite{li-etal-2024-mage}—all are diverse, high-quality, large-scale datasets. For each dataset, we randomly sampled 5,000 human-written texts as the calibration set, and 2,500 human-written texts alongside 2,500 machine-generated texts as the test set.
\paragraph{Metrics.}
We employ the FPR, defined as the proportion of human-written texts misclassified as machine-generated, as the primary metric in MGT detection. Additionally, we use the TPR ($TP@FPR$) and the F1 score ($F_1@FPR$) as metrics to evaluate detection performance. 
% Detailed hyperparameter settings are provided in Appendix~\ref{appendix: hyperparameter}.
\paragraph{Basic Detectors.}
 We selected SOTA zero-shot detectors Fast-DetectGPT \cite{bao2024fastdetectgpt} and Binoculars \cite{hans2024spotting}, as well as other zero-shot detectors including DetectGPT \cite{pmlr-v202-mitchell23a}, Likelihood, Log-Rank, and Entropy \citep{gehrmann-etal-2019-gltr, su2023detectllmleveraginglogrank, ippolito-etal-2020-automatic}. We also considered the supervised detectors OpenAI-D \cite{solaiman2019release}, which utilizes RoBERTa fine-tuned on the GPT-2 dataset.

\paragraph{Hyperparameter Settings.} Detailed hyperparameter settings and analysis refer to Appendix~\ref{appendix: hyperparameter}.
% The maximum input text length $L_\phi$ is generally determined by the tokenizer of the base detector $D_\phi$. Additionally, we used an equal-width strategy for binning the calibration set into $K$ subsets, with a typical bin width of 100 tokens. Experimental results under different bin widths are provided in Appendix F.

\subsection{False Positive Rate Constraint}
Figure~\ref{fig4} shows the FPR performance of various detectors under the MCP framework across different datasets, with $\alpha$ values selected from $\{0.2, 0.1, 0.05, 0.02, 0.01, 0.005\}$. Each subfigure corresponds to a specific $\alpha$ value, and datasets are distinguished by color. The results confirm that the FPRs are consistently constrained within the theoretical upper bound determined by $\alpha$, demonstrating MCP’s efficacy in controlling false positives. Notably, the uniformity of FPR across detectors highlights the framework's generalizability, making it an effective solution for environments demanding tight FPR control.

\begin{figure*}[!htbp]
\centering
\counterwithout{figure}{section}
\resizebox{\linewidth}{!}{
\includegraphics{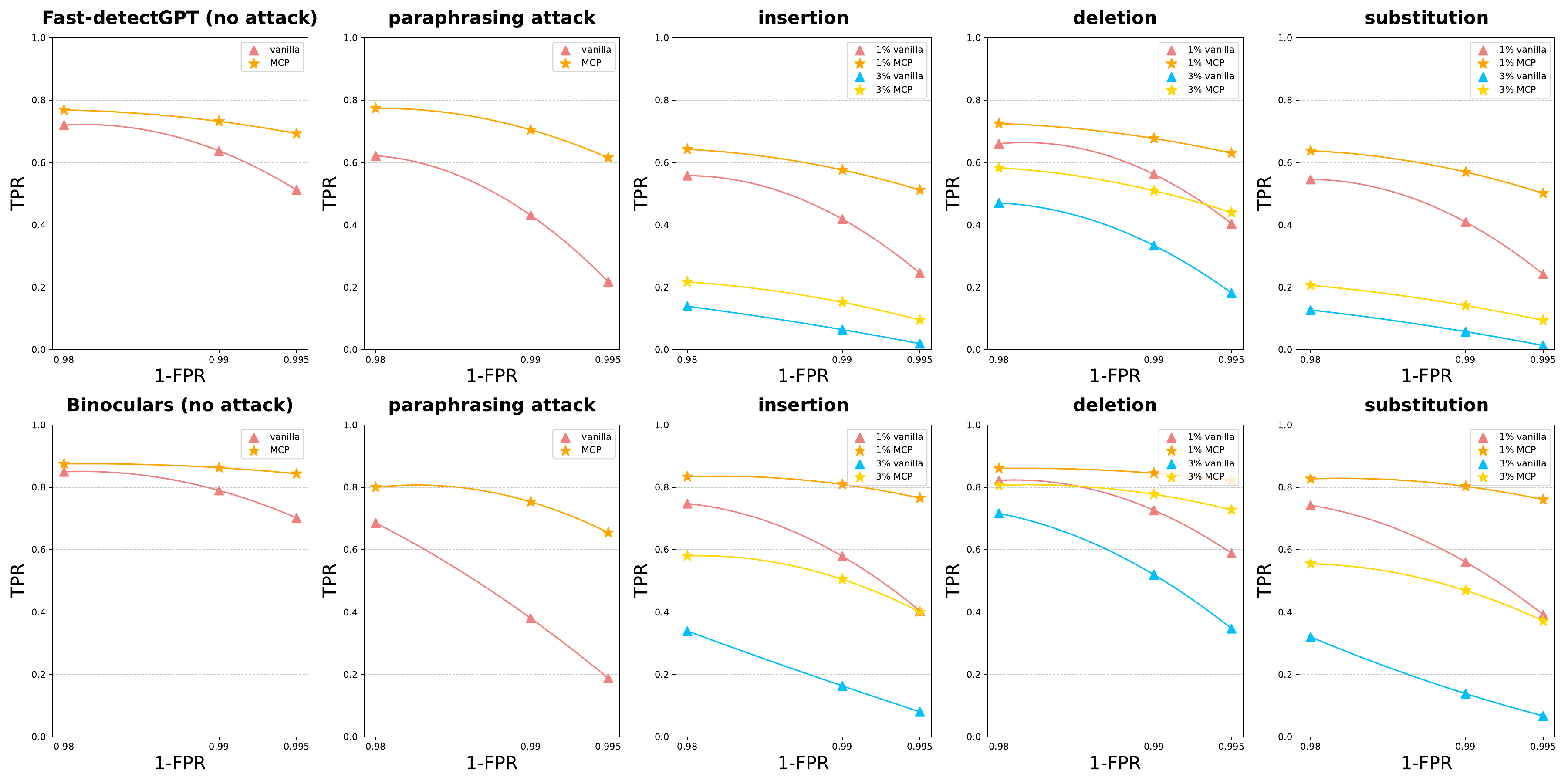}
}
\caption{Local ROC curves (with the horizontal axis representing 1 - FPR) for the basic detectors (Binoculars, Fast-DetectGPT) under different real-world attacks, both with and without the MCP framework.}
\label{fig5}
\vspace{-10pt}
\end{figure*}

\subsection{Main Results} \label{subsec: Main Results}
Table~\ref{tab:main_experiment} presents the detection results across four datasets and seven detectors. The MCP framework consistently improves detection performance compared to vanilla detectors, demonstrating strong generalizability. While the gains are modest at higher FPR levels (20\%, 10\%, and 5\%), MCP shows significant improvements at lower FPR thresholds (2\%, 1\%, and 0.5\%). On the RealDet dataset, MCP results in an average improvement of \textbf{11\%} in TPR and \textbf{8\%} in F1 score compared to vanilla detectors, and an average improvement of \textbf{10\%} in TPR and \textbf{6\%} in F1 score across all datasets. MCP enhances performance by balancing detection accuracy and FPR control through multiscaled conformal quantiles, making it well-suited for applications with strict FPR constraints.

MCP demonstrates superior performance in low-FPR scenarios. For example, on the MAGE dataset, MCP achieves relative improvements of \textbf{157\%} in $TP@0.5\%$ and \textbf{91\%} in $F1@0.5\%$. At higher FPR levels, the improvements are more limited, likely due to the proximity of multiscaled quantiles. Notably, SOTA detectors with MCP maintain high performance under stringent low-FPR constraints. On RealDet, Fast-DetectGPT reaches \textbf{69.32\%} in $TP@0.5\%$ and \textbf{81.59\%} in $F1@0.5\%$, while Binoculars achieves \textbf{84.34\%} and \textbf{91.29\%}.  
MCP's flexibility in adjusting detection thresholds through multiscaled quantiles allows for precise control of low FPRs without sacrificing performance, making it especially effective for high-precision detection.

\subsection{Ablation Study}
%Sections 5.2 and 5.3 demonstrate the superiority of MCP in constraining the FPR and enhancing detection performance. This section conducts an ablation study on the multiscaled quantiles calculation within MCP. More detailed experimental results can be found in Appendix C.
It is important to note that MCP is a framework, and the ablation study focuses on the individual modules within the framework, rather than removing the entire framework itself. Table~\ref{tab:ablation_study} compares detection performance with and without the multiscaled quantiles calculation module across different datasets. “w/o $\hat{q}_{M}$” refers to single quantile calculation based on the overall distribution. The results demonstrate that incorporating multiscaled quantiles into the MCP significantly improves detection performance. Specifically, when $\hat{q}_{M}$ is removed, the average TPR decreases by \textbf{22\%}, and the average F1 score drops by \textbf{15\%}. These findings underscore that binning the calibration set and calculating more appropriate quantiles over different length intervals enables more precise calibration, highlighting the necessity of multiscaled quantiles.

\begin{table}[t]
    \centering
    \resizebox{\linewidth}{!}{
    \large % Make the font size larger
    \begin{tabular}{lllcccc}
        \toprule
        \textbf{\textit{Dataset}} & \textbf{\textit{Detector}} & \textbf{\textit{Setting}} & \textbf{$TP@1\%$} & \textbf{$F_1@1\%$} & \textbf{$TP@0.5\%$} & \textbf{$F_1@0.5\%$} \\ 
        \midrule
        \multirow{4}{*}{MAGE}  & \multirow{2}{*}{Fast-DetectGPT}  & \textbf{MCP}  & \textbf{65.92} & \textbf{78.91} & \textbf{51.40} & \textbf{67.61} \\
                               &                        & w/o  $\hat{q}_{M}$ & 59.76 & 74.25 & 48.56 & 65.08 \\ \cmidrule(l){2-7}
                               & \multirow{2}{*}{Binoculars}     & \textbf{MCP}  & \textbf{75.80} & \textbf{85.77} & \textbf{73.32} & \textbf{84.49} \\
                               &                        & w/o  $\hat{q}_{M}$ & 50.20 & 66.49 & 24.12 & 38.72 \\
        \midrule
        \multirow{4}{*}{RealDet}  & \multirow{2}{*}{Fast-DetectGPT}  & \textbf{MCP}  & \textbf{73.20} & \textbf{83.97} & \textbf{69.32} & \textbf{81.59} \\
                               &                        & w/o  $\hat{q}_{M}$  & 64.46 & 77.91 & 51.72 & 67.95 \\\cmidrule(l){2-7}
                               & \multirow{2}{*}{Binoculars}     & \textbf{MCP}  & \textbf{86.28} & \textbf{92.28} & \textbf{84.34} & \textbf{91.29} \\
                               &                        &  w/o  $\hat{q}_{M}$  & 80.82 & 88.88 & 73.56 & 84.44 \\
        \bottomrule
    \end{tabular}
    }
    \caption{Ablation Study of multiscaled quantiles.}
    \label{tab:ablation_study}
    \vspace{-12pt}
\end{table}

\subsection{Robustness to Real-world Attacks}
%The results demonstrate that the MCP framework maintains higher TPR across all FPR thresholds, even as adversarial attacks degrade performance. Notably, MCP shows significant resilience against both paraphrasing and token-level edits, preserving robust detection capabilities compared to non-MCP detectors whose performance declines markedly under the same conditions. These results highlight the MCP's effectiveness in enhancing detection robustness under adversarial scenarios.
Figure~\ref{fig5} illustrates the robustness of the MCP framework under two types of adversarial attacks: paraphrasing and token-level edits (insertion, deletion, and substitution). Paraphrasing attacks were conducted using DIPPER \cite{DBLP:journals/corr/abs-2303-13408} to rephrase the machine-generated texts. Editing attacks involved random insertion, deletion, or substitution of tokens at rates of 1\% or 3\%.  

The results show that, under all attack scenarios, the MCP framework consistently achieves higher TPR compared to the vanilla detectors. Specifically, MCP demonstrates superior resilience, with higher true positive rates across various attack types and intensities. As attack strength increases (from 1\% to 3\%), MCP continues to maintain a more robust detection performance, whereas vanilla detectors experience more significant drops in TPR. For example, under insertion and deletion attacks, MCP outperforms the vanilla detectors by a substantial margin, indicating its effectiveness in mitigating the impact of adversarial edits. These trends highlight the effectiveness of the MCP framework in maintaining strong detection performance even in challenging adversarial scenarios, emphasizing its robustness compared to the baseline detectors.

\subsection{The Impact of Calibration Data}

\begin{figure}[t]
\centering
\counterwithout{figure}{section}
\resizebox{\linewidth}{!}{
\includegraphics{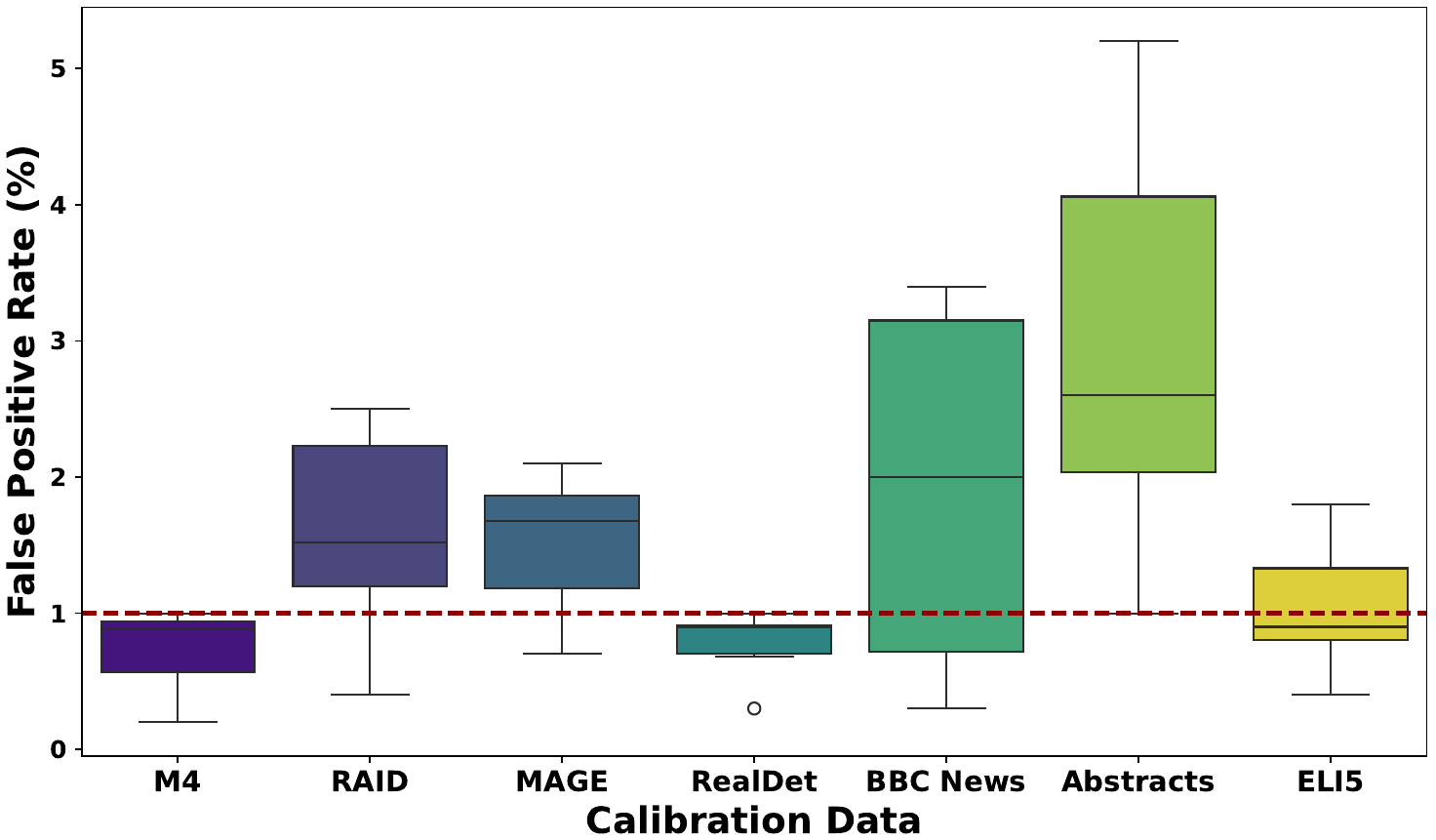}
}
\caption{FPR of Binoculars within the MCP at $\alpha = 1\%$ when calibrated with different datasets.
}
\label{fig6}
\vspace{-15pt}
\end{figure}

%\textbf{The performance of multi-domain calibration data demonstrates better results and can effectively generalize to specific domains.} Figure~\ref{fig6} shows the FPR performance of Binoculars under different calibration data. It is evident that domain-specific data (BBC News, Abstracts, ELI5) fails to achieve precise calibration, with the FPR significantly exceeding the set threshold of $\alpha$. Similarly, multi-domain data (RAID, MAGE) also fail to provide accurate calibration, which could be attributed to sampling biases in the data. In contrast, M4 and RealDet deliver better results, with RealDet demonstrating greater stability and a more significant performance improvement. Specifically, TP@1\% increased by an average of 13\% across all test sets when calibrated with RealDet.
To explore the impact of calibration data on MCP performance, we evaluate multiple datasets, including the multi-domain datasets (M4, RAID, MAGE, RealDet), as well as domain-specific datasets such as news writing (BBC News), academic writing (Abstracts), and social media text (ELI5). As shown in Figure~\ref{fig6}, we used a single dataset for calibration, while the remaining datasets were sequentially used as test data.

\textbf{The use of diverse calibration data generally leads to better performance and improved generalization.} Figure~\ref{fig6} shows the FPR of MCP under different calibration data. The results demonstrate that domain-specific datasets (BBC News, Abstracts, ELI5) lead to relatively higher FPRs, suggesting that these datasets are less effective in achieving precise calibration. In contrast, multi-domain datasets (RAID, MAGE) generally perform better, although they still exhibit some limitations due to inherent biases in the data. Notably, M4 and RealDet calibration both yield promising results, with RealDet providing slightly more stable and consistent improvements. Specifically, TP@1\% increased by an average of 13\% across all test sets when calibrated with RealDet. However, M4 also produces competitive calibration results, highlighting that while RealDet offers a slight edge, diverse calibration data from different domains still plays a critical role in enhancing performance without being overly reliant on a single dataset.

% Obtaining human-written texts for calibration from unseen sources is often impractical. To address this challenge, we propose the RealDet dataset, which provides a fixed and well-distributed calibration set. Figure~\ref{fig6} demonstrates the importance of using such a calibration set. In the experiment, texts sampled from different datasets are used as calibration and test sets to simulate detection under unseen text sources. The results show that when the calibration set is sourced from the RealDet dataset, the detector consistently maintains an effective FPR. Notably, $TP@1\%$ also improved by an average of \textbf{13\%} across all test sets, underscoring the importance of a robust, pre-defined calibration set in ensuring consistent detection performance in real-world scenarios.
\subsection{Comparison with Other Calibration Methods}
We compared MCP with other calibration methods in Appendix \ref{appendix: other calibration mehods}, including metric-based maximizing F1 \cite{lipton2014thresholdingclassifiersmaximizef1}, probability distribution-based Platt Scaling \cite{Platt1999ProbabilisticOF} and Isotonic Regression\cite{Brunk1973StatisticalIU}. While these methods offer a modest improvement in detection, they fall short of effectively controlling the FPR. In contrast, MCP achieves SOTA classification performance while maintaining an exceptionally low FPR, ensuring higher reliability.

\section{Related Work}
\paragraph{MGT Detection.}
Existing detectors can be broadly categorized into two main types: zero-shot detectors and supervised detectors. (1) Zero-shot detectors leverage statistical measures extracted by LLMs to identify outliers \citep{gehrmann-etal-2019-gltr, su2023detectllmleveraginglogrank, ippolito-etal-2020-automatic, yang2023dnagptdivergentngramanalysis}. For instance, the impressive DetectGPT \cite{pmlr-v202-mitchell23a}, based on the assumption that MGT is more likely to lie at a local optimum of the log probability, compares log probabilities across multiple perturbations to detect MGT. Fast-DetectGPT \cite{bao2024fastdetectgpt} further improves the text perturbation process of its predecessor, significantly enhancing detection efficiency. Binoculars \cite{hans2024spotting} uses cross perplexity between two models from different perspectives to address poor performance when detecting high-perplexity text. (2) supervised detectors typically train a classification model using human-written and machine-generated texts \citep{solaiman2019releasestrategiessocialimpacts, uchendu-etal-2020-authorship, Fagni_2021, zhang2024detecting, tian2024multiscale, Pu2022DeepfakeTD, DBLP:journals/corr/abs-2307-03838, DBLP:conf/ndss/KumariPFS24}. Specifically, OpenAI-D \cite{solaiman2019release} fine-tuned a RoBERTa model on GPT-2 generated text to detect MGT.

\paragraph{MGT Detection Datset.}
Turing Bench \cite{uchendu-etal-2021-turingbench-benchmark} collected 200k human-written texts and machine-generated texts from 19 different models. However, it has become outdated due to the less advanced models. Subsequently, researchers constructed datasets focusing on specific advanced models or particular domains \citep{Fagni_2021, DBLP:journals/corr/abs-2304-12008, mosca-etal-2023-distinguishing}. For instance, \citet{guo2023closechatgpthumanexperts} built the HC3 dataset by collecting nearly 40k questions covering multiple domains along with corresponding answers generated by human experts and ChatGPT. More recent efforts have introduced large-scale, cross-domain, and cross-model benchmarks, such as the MGTBench \cite{he2024mgtbenchbenchmarkingmachinegeneratedtext}, M4 \cite{wang-etal-2024-m4}, MAGE \cite{li-etal-2024-mage}, RAID \cite{dugan-etal-2024-raid} and DetectRL \cite{wu2024detectrl} datasets.

\section{Conclusion}
In this paper, we introduce a reliable machine-generated text detection framework via multiscaled conformal prediction (MCP), which constrains FPRs to mitigate potential societal harms while simultaneously enhancing detection performance. Extensive experiments across seven detectors and four datasets validate the effectiveness of MCP and demonstrate its ability to improve robustness. In future work, we will continue to update our high-quality dataset, RealDet, to address the challenges posed by the rapidly evolving LLMs. Additionally, we plan to investigate an advanced detector, aiming to maintain exceptional detection performance even under stringent FPR within MCP.

\section*{Limitations}
In our experiments, we found that although we applied a multiscaled optimization strategy using fixed-width binning based on CP, different bin widths consistently corresponded to varying detection performance. Therefore, a more flexible binning strategy could potentially lead to better detection results, an area we have not explored in depth.

\section*{Ethics Statement}
Detection inherently carries an accusatory implication. While our work constrains the upper bound of the false positive rate, offering more reliable insights to users, we strongly oppose using the detection results from this framework as direct evidence in any punitive context. Regardless of the accuracy of the detection, such use could cause significant harm. Additionally, per the Code of Ethics, no private data or non-public information was used in constructing our dataset.

%\bibliography{anthology,ACL2025}
% \bibliographystyle{acl_natbib}

\appendix
\begin{figure*}[htbp]
\centering
\counterwithout{figure}{section}
\resizebox{\linewidth}{!}{
\includegraphics{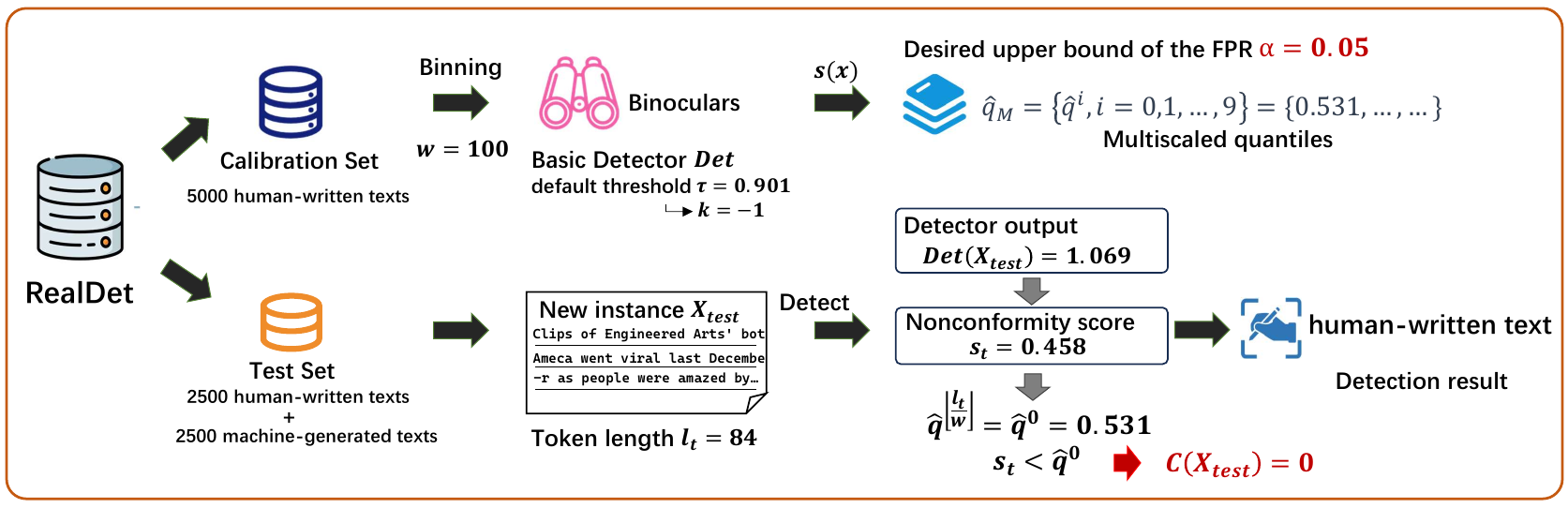}
}
\caption{Detailed Detection Process of a Single Instance within the MCP Framework.}
\label{fig: case_study}
\end{figure*}
\section{Case Study}\label{appendix: case study}

Figure \ref{fig: case_study} illustrates the detailed detection steps of a single instance within the MCP framework when using Binoculars as the detector, with 
$w=100$ and $\alpha=0.05$. First, the multiscaled quantiles $\hat{q}_M$ are calculated based on the nonconformity score distribution of the calibration data. Next, for a given instance $X_\text{test}$, its token length is determined and processed through Binoculars to obtain the output $Det(X_\text{test})$. The output $Det(X_\text{test}$) is then converted into a nonconformity score $s_t$
following Equation \ref{eq:8}. Subsequently, using $l_t$ and $w$, the corresponding quantile $\hat{q}^0$ from $\hat{q}_M$ is retrieved for the matching length interval. Finally, the prediction result $\mathcal{C}(X_\text{test})$ is determined based on the comparison between $s_t$ and $\hat{q}^0$.

\section{Corollary Proof}\label{appendix: proof}

This section provides a detailed proof of Corollary~\ref{corollary 1}.

\begin{proof}

\textbf{Assumptions:}
\begin{enumerate}
    \item The calibration set $D_{cal}$ and the test set $D_{test}$ are independent and identically distributed (i.i.d.).
    \item The conformal prediction framework is employed to calibrate a decision rule based on the calibration set $D_{cal}$, which is then applied to the test set $D_{test}$.
    \item The nonconformity scores \( s(x) \) are properly defined such that higher scores indicate a lower likelihood of the text being HWT.
\end{enumerate}

Within the MCP framework, all of the assumptions above are satisfied.

\textbf{Objective:}  
To demonstrate that under the MCP framework, the False Positive Rate (FPR) on the test set $D_{test}$ does not exceed the predefined threshold $\alpha$.

\textbf{Proof Steps:}

\textbf{Nonconformity Scores Assignment:}  
Assign a nonconformity score \( s(x) \) to each instance \( x \) in both $D_{cal}$ and $D_{test}$. These scores quantify how atypical an instance is with respect to the detector.

\textbf{Multiscaled Quantiles Calculation:}
\begin{enumerate}
    \item Bin the calibration set and sort the nonconformity scores of the subset \( D^i_{cal} \) in ascending order:
    \[
    s^i_{(1)} \leq s^i_{(2)} \leq \dots \leq s^i_{(n)}
    \]
    \item By applying Equation~\ref{Eq: quantile} to compute $\hat{q}_{M}$, the following condition is guaranteed:
    \[
    P(s^i(x) > q^i) \leq \alpha \quad \text{for} \quad x \in D^i_{cal}
    \]
\end{enumerate}

\textbf{Bounding the False Positive Rate:}
\begin{enumerate}
    \item Since $D_{cal}$ and $D_{test}$ are i.i.d., the distribution of nonconformity scores in $D_{cal}$ mirrors that of $D_{test}$. After binning based on text length, and since the selection is performed solely on individual attributes, the subsets ($D^i_{cal}$ and $D^i_{test}$) obtained for different length intervals remain i.i.d.
    
    \item Consequently, the same $\hat{q}_{M}$, when applied to the test set, continues to satisfy the following condition:
    \[
    P(s^i(x) > q^i) \leq \alpha \quad \text{for} \quad x \in D^i_{test}
    \]
    
    \item The FPR is computed from the instances across all length intervals in $D_{test}$:
    \[
    % \small
    \text{FPR} = \mathbb{E}\left(P\left(s^i(x) > q^i \mid i=1, 2, \ldots\right)\right) \leq \alpha
    \]
\end{enumerate}

\end{proof}

\section{Details of RealDet}
In this subsection, we will provide additional detailed information regarding various aspects of RealDet.

\subsection{Specific Quantities in Data Sources}\label{appendix: data sources}
Table~\ref{RealDet-English} presents the detailed distribution of English text data in the RealDet dataset, categorized by their sources. The dataset contains over 100k human-written texts, and more than 618k machine-generated texts, culminating in a total of 718k English texts.  Table~\ref{RealDet-Chinese} illustrates the sources and specific quantities of Chinese text data in the RealDet dataset. The Chinese text data is constructed with reference to the HC3 dataset sources \cite{guo2023closechatgpthumanexperts}, resulting in a total exceeding 129k texts.

\begin{table}[ht]
\centering
\resizebox{\columnwidth}{!}{
\begin{tabular}{l|r|r|r}
\hline
\textbf{Dataset}       & \textbf{HWT} & \textbf{MGT} & \textbf{Total} \\ \hline
ELI5         & 17,111              & 324,220               & 341,331        \\ 
WikiQA        & 1,187               & 23,740                & 24,927         \\ 
Wikipedia     & 5,612               & 24,840                & 33,682         \\ 
Medical Dialog& 1,248               & 24,960                & 26,208         \\ 
FiQA         & 3,933               & 78,660                & 82,593         \\ 
Xsum          & 8,000               & 9,600                 & 17,600         \\ 
TLDR          & 7,887               & 9,600                 & 17,487         \\ 
BBC News      & 2,225               & 8,900                 & 11,125         \\ 
WritingPrompt & 8,000               & 9,600                 & 17,600         \\ 
ROC Stories   & 5,000               & 55,941               & 52,941         \\ 
Yelp          & 8,000               & 9,600                 & 17,600         \\ 
IMDB          & 8,000               & 9,600                 & 17,600         \\ 
CMV           & 8,000               & 9,600                 & 17,600         \\ 
Abstracts     & 8,000               & 9,600                 & 17,600         \\ 
SQuAD         & 8,000               & 9,600                 & 17,600         \\ \hline
\textbf{Total}         & 100,203            & 618,061            & 718,264        \\ \hline
\end{tabular}
}
\caption{Specific Quantities in RealDet-English.}
\label{RealDet-English}
\end{table}

\begin{table}[ht]
\centering
\resizebox{\columnwidth}{!}{  % Resizing the table to fit within the column width
\begin{tabular}{l|r|r|r}
\hline
\textbf{Dataset}       & \textbf{HWT} & \textbf{MGT} & \textbf{Total} \\ \hline
\parbox[t]{4.5cm}{WebTextQA \& BaikeQA \\ \cite{bright_xu_2019_3402023}} & 3,293 & 29,637 & 32,930 \\
\parbox[t]{5cm}{BaiduBaike \\ \url{https://baike.baidu.com/}} & 4,600 & 41,685 & 46,285 \\
\parbox[t]{5cm}{NLPCC-DBQA \\ \cite{DBLP:conf/nlpcc/Duan16}} & 1,709 & 16,381 & 18,090 \\
\parbox[t]{5cm}{Medical Dialog \\ \cite{he2020meddialoglargescalemedicaldialogue}} & 1,074 & 9,666 & 10,740 \\
\parbox[t]{5cm}{FinanceZhidao \\ \href{https://github.com/SophonPlus/ChineseNlpCorpus}{(Finance Zhidao)}} & 689 & 6,201 & 6,890 \\
\parbox[t]{5cm}{Chinese Psychological QA \\ \href{https://aistudio.baidu.com/aistudio/datasetdetail/38489}{(Chinese Psychological QA)}} & 1,099 & 9,891 & 10,990 \\
\parbox[t]{5cm}{LegalQA \\ \href{https://github.com/siatnlp/LegalQA}{(Legal QA)}} & 372 & 3,348 & 3,720 \\ \hline
\textbf{Total} & 12,836 & 116,809 & 129,645 \\ \hline
\end{tabular}
}
\caption{Data Sources and Specific Quantities in RealDet-Chinese.}
\label{RealDet-Chinese}
\end{table}

\subsection{Model Supplement}
\label{appendix: model supplement}
Table~\ref{RealDet-Model} presents all LLMs along with their corresponding text quantities. In collecting machine-generated texts, we considered two distinct settings: black-box models and white-box models. From the black-box models, we selected 9 popular and powerful LLMs for data collection, namely Deepseek-R1 \cite{deepseekai2025deepseekr1incentivizingreasoningcapability}, GPT-4, ChatGPT, PaLM 2, Ernie Bot 3.5 Turbo, Spark Desk 2.0, Qwen Turbo, 360GPT S2 V9, and Minimax Abab 5.5. Similarly, from the white-box models, we selected 13 LLMs, including LLaMA2-13B \cite{touvron2023llama2openfoundation}, ChatGLM2-6B \cite{glm2024chatglm}, MOSS-moon-003, MPT-7B \cite{MosaicML2023Introducing}, InternLM-7B \cite{cai2024internlm2}, Alpaca-7B \cite{alpaca}, Guanaco-7B \cite{dettmers2023qloraefficientfinetuningquantized}, Vicuna-13B \cite{vicuna2023}, BLOOMz-7B \cite{muennighoff2023crosslingualgeneralizationmultitaskfinetuning}, Falcon-7B \cite{refinedweb}, OPT-6.7B \cite{zhang2022opt}, Baichuan-13B, and Flan-T5-XXL \cite{https://doi.org/10.48550/arxiv.2210.11416}. The number of machine-generated English texts exceeds 618k, while for Chinese text generation models, there are nine models with over 116k texts generated. The total number of texts exceeds 847k.

\begin{table}[ht]
\centering
\resizebox{\columnwidth}{!}{
\begin{tabular}{l|r|r|r}
\hline
\textbf{Source Model}       & \textbf{En-Text} & \textbf{Cn-Text} & \textbf{Total} \\ \hline
Deepseek-R1          & 12,800              & 5,000               & 17,800        \\
GPT-4          & 14,400              & -               & 14,400        \\ 
ChatGPT        & 23,861               & 12,828                & 36,689         \\ 
PaLM2     & 22,255               & -                & 22,255         \\ 
Ernie Bot turbo 3.5& 46,314               & 12,851                & 59,165         \\ 
Spark Desk 2.0          & 45,028               & 9,076                & 54,104         \\ 
Qwen turbo          & 46,359               & 12,851                & 59,210         \\ 
360GPT S2 V9          & 45,929              & 12,851                 & 58,780         \\ 
Minimax abab 5.5      & 20,119               & -                 & 20,119         \\ \hline
LLaMA2-13B & 33,489               & -                 & 33,489         \\ 
ChatGLM2-6B   & 29,251               & 12,800               & 42,051         \\ 
MOSS-moon-003          & 29,226              & 12,851                & 42,077         \\ 
MPT-7B          & 29,315               & -                 & 29,315         \\ 
InterLM-7B           & 23,573               & -                 & 23,573         \\ 
Alpaca-7B     & 28,953               & -                 & 28,953         \\ 
Guanaco-7B     & 29,264               & -                 & 29,264         \\ 
Vicuna-13B     & 29,047               & -                 & 29,047         \\ 
BLOOMz-7B     & 7,269               & 12,850                 & 20,119         \\ 
Falcon-7B     & 23,331               & -                 & 23,331         \\ 
OPT-6.7B     & 27,810               & -                 & 27,810         \\ 
Baichuan-13B     & 29,259               & 12,851                 & 42,110         \\ 
Flan-T5-XXL     & 21,129               & -                 & 21,129         \\ \hline
Human         & 100,203            & 12,836            & 113,039        \\ \hline
\textbf{Total}         & 718,264            & 129,645            & 847,909        \\ \hline
\end{tabular}
}
\caption{Specific Quantities in different LLMs generated texts.}
\label{RealDet-Model}
\end{table}

\subsection{Prompt Design}
\label{appendix: prompt design}
In this study, we designed 3 types of generic prompts—continuation writing,  topical writing, and question-answering—for LLMs to generate texts across 15 distinct domains. Table~\ref{RealDet-Prompt1} presents sample prompts for the continuation type, Table~\ref{RealDet-Prompt2} showcases examples of thematic writing prompts, and Table~\ref{RealDet-Prompt3} illustrates examples of question-answering prompts.

\begin{figure*}[t]
\centering
\counterwithout{figure}{section}
\resizebox{\linewidth}{!}{
\includegraphics{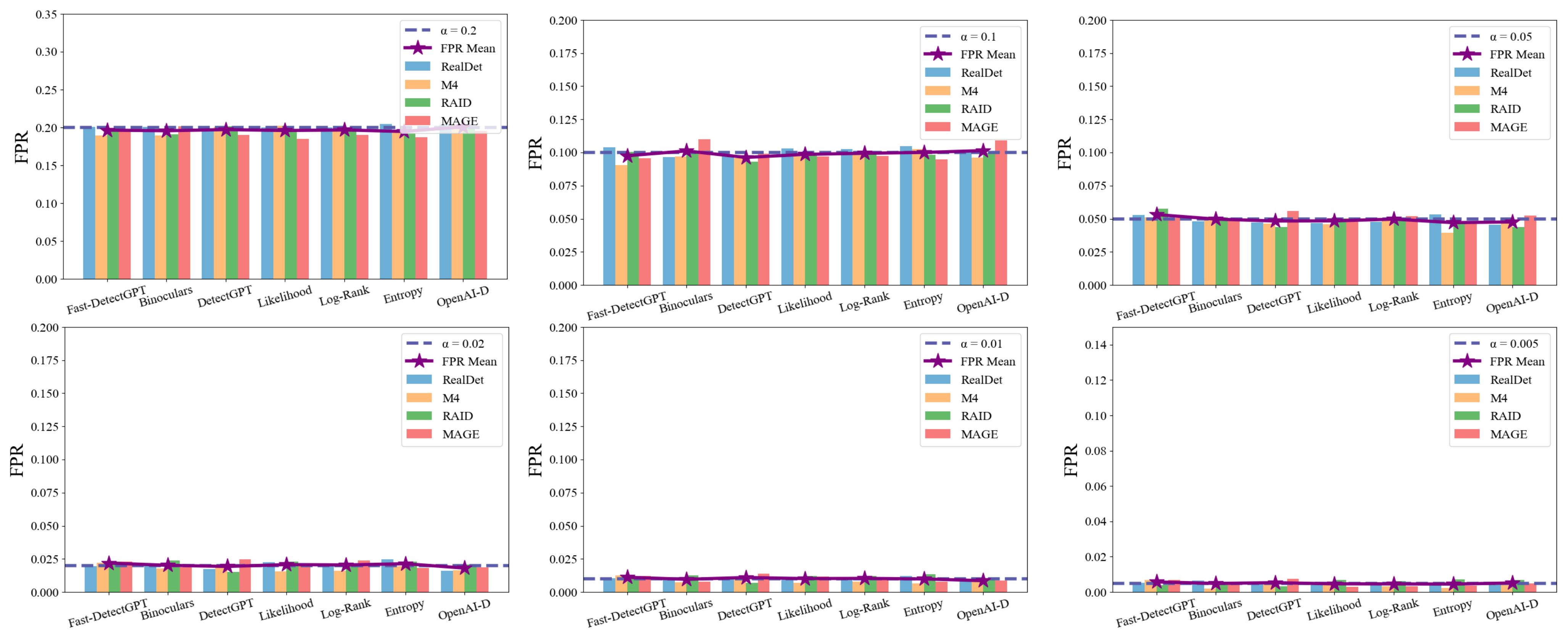}
}
\caption{The FPR of various detectors within Traditional CP framework across all datasets, after applying alpha constraints with values of alpha set to 0.2, 0.1, 0.05, 0.02, 0.01, and 0.005.}
\label{CP_FPR}
\end{figure*}

\begin{figure*}[t]
\centering
\counterwithout{figure}{section}
\resizebox{\linewidth}{!}{
\includegraphics{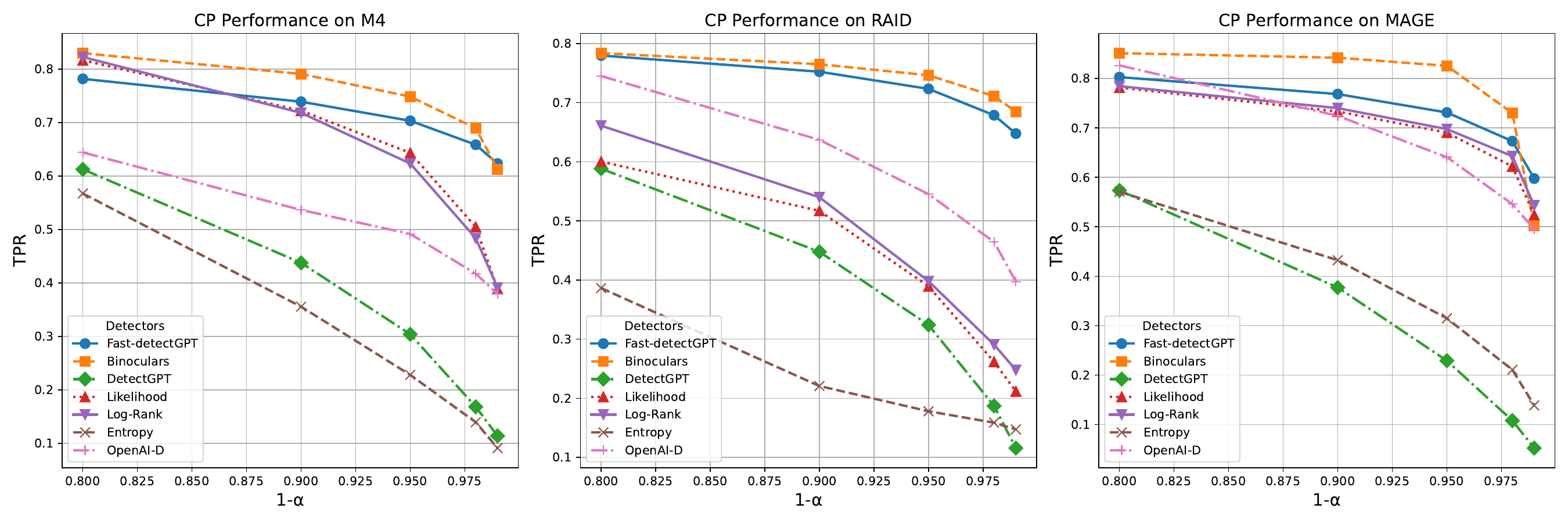}
}
\caption{The TPR of various detectors within Traditional CP framework across 3 datasets.}
\label{CP_TPR}
\end{figure*}

\subsection{Adversarial Attacks}

In this study, we do not consider adversarial attacks on human-written texts, as evading detection with human-written content is deemed inconsequential. Instead, we focus on adversarial attacks in the context of machine-generated texts by introducing 2 common attack types: paraphrasing attacks and editing attacks. For paraphrasing attacks, we employ DIPPER with hyperparameters set to a lexical diversity of 60 and a syntactic diversity of 60. This level of paraphrasing is sufficient to potentially bypass state-of-the-art (SOTA) detectors. Regarding editing attacks, we utilize the GPT-2 tokenizer to encode the text and obtain a token sequence. We then apply random insertions, deletions, and substitutions to the token sequence at proportions of 1\%, 3\%, and 5\%. The tokens inserted and substituted are randomly selected from the tokenizer's vocabulary. In the adversarial attack scenario, each raw text is associated with 10 adversarial texts (calculated as 1 original + 3 proportions × 3 types of edits).

\section{Performance of the Traditional CP}

Figure~\ref{CP_FPR} shows the FPR performance of different detectors under the traditional CP framework across various datasets. We observe that the traditional CP framework effectively constrains the upper bound of the FPR. Figure~\ref{CP_TPR} presents the TPR performance of different detectors within the traditional CP framework. Although the traditional CP framework successfully limits the FPR upper bound, we find that the TPR sharply decreases under low FPR settings. Even SOTA detectors experience a significant decline in TPR, making it difficult to detect MGTs. This observation serves as the motivation for our proposed MCP approach.

\section{Main Experiment Supplement}
Tables~\ref{tab:main_experiment_M4},~\ref{tab:main_experiment_RAID}, and~\ref{tab:main_experiment_MAGE} present a comparative analysis of the detection performance of 7 different detectors on the M4, RAID, and MAGE datasets, respectively, both with and without the MCP framework. These tables serve as supplementary material to the primary experiments. The results consistently demonstrate that the MCP framework enhances the detection capabilities of the detectors, particularly under settings with low false positive rates. Furthermore, it was observed that non-state-of-the-art (non-SOTA) detectors sometimes do not exhibit performance improvements when integrated with the MCP framework, especially when their baseline performance is bad. We attribute this phenomenon to the possibility that inaccurate outputs from non-SOTA detectors may hinder the calibration process within the MCP framework. This observation aligns with the conclusions drawn in Subsection~\ref{subsec: Main Results}, where it was noted that accurate outputs from SOTA detectors facilitate more effective calibration within MCP.

\begin{figure*}[!htbp]
\centering
\counterwithout{figure}{section}
\resizebox{\linewidth}{!}{
\includegraphics{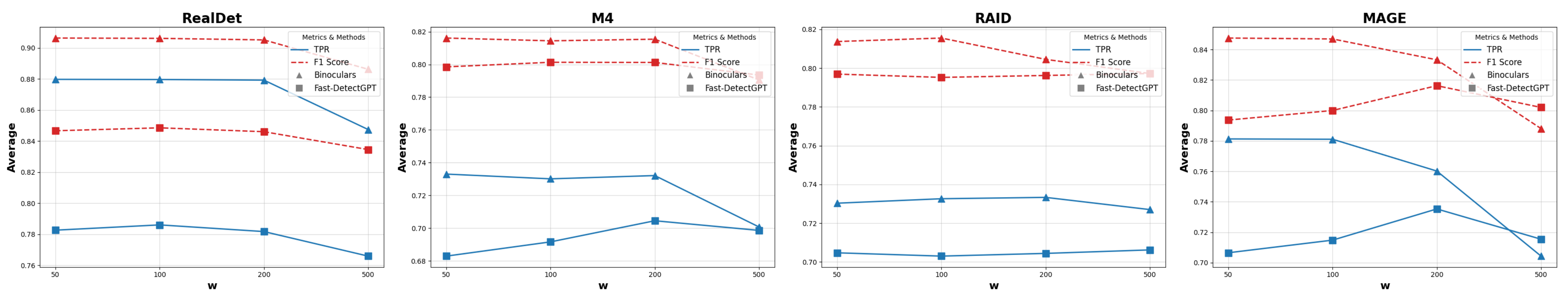}
}
\caption{Hyperparameter Sensitivity Analysis of $w$.}
\label{hyper_analysis}
\end{figure*}

\section{Hyperparameter Sensitivity Analysis} \label{appendix: hyperparameter}
The MCP framework encompasses 3 hyperparameters: the upper bound on the FPR ($\alpha$), the maximum input length ($L_{\text{max}}$), and the bin width ($w$). In the main experiments, the upper bound
$\alpha$ is typically selected from the set $\{0.2, 0.1, 0.05, 0.02, 0.01, 0.005\}$. $L_{\text{max}}$ is contingent upon the basic detector integrated within the framework. For instance, the OpenAI-D detector, which is fine-tuned based on RoBERTa, utilizes $L_{\text{max}} = 512$,  whereas Fast-DetectGPT, which employs GPT-2 for sampling and scoring, adopts $L_{\text{max}} = 1024$. The bin width $w$ is influenced by both the data distribution and the output distribution. In the main experiments, a default bin width of $w=100$ is employed, as this width yields the most optimal and stable performance within the framework.

Figure~\ref{hyper_analysis} presents the performance of varying bin width ($w$) values across different datasets and detectors. We observe that performance remains consistently strong when $w \leq 200$. However, at $w=500$, detection performance noticeably declines, indicating that excessively large bin widths undermine the MCP framework's ability to enhance detection performance.

\textbf{The optimal bin width is influenced by the distribution of the textual data.} Specifically, within the M4 dataset, a bin width of $w=200$ consistently outperforms other values in detection performance. We attribute this to the inherent text distribution of the M4 dataset, where the majority of texts pertain to academic writing and peer review, introducing a certain degree of bias.

\textbf{The distribution of the detector's outputs affects the optimal selection of $w$}. The original output distribution of Binoculars is more concentrated, leading to a similarly concentrated distribution of the transformed nonconformity scores. Consequently, a smaller and more refined bin width enhances detection performance. In contrast, Fast-DetectGPT exhibits a relatively dispersed and smooth output distribution, suggesting that a larger bin width may achieve superior detection performance in this context.

\begin{table}[t]
    \centering
    \resizebox{\linewidth}{!}{
    \large % Make the font size larger
    \begin{tabular}{lccc}
        \toprule
        \textbf{\textit{Detector}} & \textbf{\textit{Method}} & \textbf{$FPR$} & \textbf{$F_1$}  \\ 
        \midrule
        \multirow{5}{*}{Fast-DetectGPT}  & \textbf{vanilla}  & 9.84 & 82.92 \\
                               & \textbf{Maximizing $F_1$} & 9.24 & 83.22  \\ 
                               & \textbf{Platt Scaling}  & 8.46 & 83.38 \\
                               & \textbf{Isotonic Regression}  & 8.78 & 83.12 \\
                               & \textbf{MCP}  & \textbf{1.98} & \textbf{85.90} \\
                               
        \midrule
        \multirow{5}{*}{Binoculars}  & \textbf{vanilla}  & 7.68 & 89.90 \\
                               & \textbf{Maximizing $F_1$} & 7.26 & 90.12  \\ 
                               & \textbf{Platt Scaling}  & 6.54 & 90.18 \\
                               & \textbf{Isotonic Regression}  & 7.34 & 89.98 \\
                               & \textbf{MCP}  & \textbf{0.72} & \textbf{92.28} \\
        \bottomrule
    \end{tabular}
    }
    \caption{Detection Performance with Different Calibration Methods.}
    \label{tab: other calibration methods}
\end{table}

\section{Comparison with Other Calibration Methods} \label{appendix: other calibration mehods}
Existing calibration methods can be categorized into two main types: metric-based methods (e.g., maximizing F1) and probability distribution-based methods (e.g., Platt Scaling and Isotonic Regression). Maximizing F1 derives the optimal threshold by identifying the threshold that maximizes the F1 score, whereas Platt Scaling and Isotonic Regression modify the output probability distribution to enhance performance. MCP differs from other calibration methods in the following two key aspects: (1) MCP effectively constrains the upper bound of the FPR, whereas other calibration methods primarily optimize classification performance without explicitly controlling FPR. (2) MCP is a zero-shot, dynamic threshold calibration method. Maximizing F1 determines a fixed threshold that maximizes F1 performance, while Platt Scaling and Isotonic Regression train a calibration model to optimize the output probability distribution and then classify instances based on the newly calibrated probabilities. In contrast, MCP requires no additional training and dynamically adjusts the threshold across different text lengths, demonstrating its efficiency and flexibility.

Table \ref{tab: other calibration methods} compares the experimental performance of MCP with other calibration methods. Using the RealDet dataset, we conducted experiments on Fast-DetectGPT and Binoculars with different calibration methods. The results indicate that although other methods (maximizing F1, Platt Scaling, and Isotonic Regression) achieve modest improvements in classification performance, they still exhibit dangerously high FPRs (FPR > 5\%). In contrast, MCP achieves SOTA classification performance and maintains an exceptionally low FPR (FPR < 2\%), ensuring higher reliability.

\begin{table*}[!t]
    \centering
    \resizebox{\textwidth}{!}{
    \large
    \begin{tabular}{llcccccccccccc}
        \toprule
        \textbf{\textit{Detector}} & \textbf{\textit{Algorithm}} & \textbf{$TP@20\%$} & \textbf{$F_1@20\%$} & \textbf{$TP@10\%$} & \textbf{$F_1@10\%$} & \textbf{$TP@5\%$} & \textbf{$F_1@5\%$} & \textbf{$TP@2\%$} & \textbf{$F_1@2\%$} & \textbf{$TP@1\%$} & \textbf{$F_1@1\%$} & \textbf{$TP@0.5\%$} & \textbf{$F_1@0.5\%$} \\ 
        \midrule
        \rowcolor[gray]{0.9} \multicolumn{14}{c}{\textbf{M4}} \\
        \multirow[t]{3}{*}{Likelihood} & vanilla & 81.48 & 80.87 & 72.76 & 79.62 & 65.12 & 76.54 & 54.72 & 69.85 & 43.92 & 60.61 & 35.96 & 52.70 \\
                                  & \textbf{MCP} & 87.28 & 84.17 & 78.68 & 83.52 & 70.04 & 80.10 & 58.12 & 72.89 & 48.28 & 64.88 & 36.04 & 52.83 \\\midrule
        \multirow[t]{3}{*}{Log-Rank} & vanilla & 82.52 & 81.49 & 71.84 & 79.03 & 62.04 & 74.28 & 50.76 & 66.47 & 42.96 & 59.68 & 35.32 & 52.02 \\
                                  & \textbf{MCP} & 87.24 & 84.32 & 78.80 & 83.72 & 70.12 & 80.12 & 56.68 & 71.66 & 48.48 & 64.99 & 38.56 & 55.51 \\\midrule
        \multirow[t]{3}{*}{Entropy} & vanilla & 57.12 & 64.50 & 35.28 & 48.53 & 25.12 & 38.61 & 14.00 & 24.14 & 10.48 & 18.80 & 8.48 & 15.57 \\
                                  & \textbf{MCP} & 64.80 & 70.59 & 45.48 & 58.71 & 28.44 & 42.82 & 14.84 & 25.53 & 9.88 & 17.87 & 6.48 & 12.14 \\\midrule
        \multirow[t]{3}{*}{DetectGPT} & vanilla & 61.20 & 67.53 & 44.68 & 57.77 & 31.12 & 45.72 & 16.96 & 28.51 & 10.08 & 18.15 & 7.28 & 13.51 \\
                                  & \textbf{MCP} & 62.00 & 68.19 & 46.28 & 59.35 & 31.52 & 46.12 & 18.36 & 30.53 & 11.52 & 20.49 & 7.08 & 13.17 \\\midrule
        \multirow[t]{3}{*}{OpenAI-D} & vanilla & 64.00 & 69.53 & 54.12 & 65.95 & 49.60 & 64.17 & 42.60 & 58.92 & 38.56 & 55.26 & 35.88 & 52.63\\
                                  & \textbf{MCP} & 67.24 & 71.43 & 59.28 & 69.84 & 52.08 & 66.45 & 45.04 & 61.28 & 38.92 & 55.73 & 35.04 & 51.76 \\\midrule
        \multirow[t]{3}{*}{Fast-DetectGPT} & vanilla & 78.56 & 79.13 & 74.44 & 80.72 & 70.04 & 80.03 & 65.32 & 78.08 & 60.56 & 74.97 & 54.84 & 70.60 \\
                                  & \textbf{MCP}  & \textbf{79.56} & \textbf{80.35} & \textbf{75.24} & \textbf{81.50} & \textbf{71.72} & \textbf{81.33} & \textbf{67.36} & \textbf{79.55} & \textbf{62.44} & \textbf{76.33} & \textbf{58.64} & \textbf{73.67} \\
                                  \midrule
        \multirow[t]{3}{*}{Binoculars} & vanilla & 83.20 & 81.87 & 79.44 & 83.87 & 74.80 & 83.19 & 69.72 & 81.21 & 64.52 & 77.96 & 55.36 & 71.03 \\
                                  & \textbf{MCP} & 83.08 & \textbf{82.03} & 78.84 & \textbf{83.96} & \textbf{75.72} & \textbf{83.87} & \textbf{71.20} & \textbf{82.27} & \textbf{66.68} & \textbf{79.59} & \textbf{62.72} & \textbf{76.88} \\
        \bottomrule
    \end{tabular}
    }
    \caption{Main Experimental Supplement on the \textbf{M4} dataset. “Vanilla” refers to the detector's original configuration, whereas “MCP” denotes detectors with MCP framework.}
    \label{tab:main_experiment_M4}
\end{table*}

\begin{table*}[!t]
    \centering
    \resizebox{\textwidth}{!}{
    \large
    \begin{tabular}{llcccccccccccc}
        \toprule
        \textbf{\textit{Detector}} & \textbf{\textit{Algorithm}} & \textbf{$TP@20\%$} & \textbf{$F_1@20\%$} & \textbf{$TP@10\%$} & \textbf{$F_1@10\%$} & \textbf{$TP@5\%$} & \textbf{$F_1@5\%$} & \textbf{$TP@2\%$} & \textbf{$F_1@2\%$} & \textbf{$TP@1\%$} & \textbf{$F_1@1\%$} & \textbf{$TP@0.5\%$} & \textbf{$F_1@0.5\%$} \\ 
        \midrule
        \rowcolor[gray]{0.9} \multicolumn{14}{c}{\textbf{RAID}} \\
        \multirow[t]{3}{*}{Likelihood} & vanilla & 64.37 & 69.82 & 52.47 & 64.59 & 38.70 & 53.86 & 24.80 & 39.12 & 19.37 & 32.18 & 14.00 & 24.46 \\
                                  & \textbf{MCP} & 62.93 & 68.87 & 53.57 & 65.59 & 44.90 & 60.00 & 36.50 & 52.63 & 32.17 & 48.20 & 26.63 & 41.83 \\\midrule
        \multirow[t]{3}{*}{Log-Rank} & vanilla & 66.10 & 71.01 & 54.10 & 65.91 & 40.63 & 55.80 & 28.83 & 44.08 & 22.67 & 36.66 & 19.77 & 32.88 \\
                                  & \textbf{MCP} & 64.97 & 70.26 & 54.73 & 66.60 & 46.47 & 61.61 & 39.43 & 55.70 & 34.23 & 50.59 & 29.40 & 45.23 \\\midrule
        \multirow[t]{3}{*}{Entropy} & vanilla & 40.03 & 50.06 & 22.37 & 33.80 & 17.90 & 29.14 & 15.57 & 26.48 & 14.13 & 24.54 & 11.77 & 20.96 \\
                                  & \textbf{MCP} & 42.33 & 52.30 & 29.33 & 42.37 & 23.53 & 36.61 & 17.43 & 29.11 & 14.73 & 25.43 & 12.27 & 21.74 \\\midrule
        \multirow[t]{3}{*}{DetectGPT} & vanilla & 58.57 & 65.62 & 46.53 & 59.45 & 34.50 & 49.46 & 21.20 & 34.41 & 14.67 & 25.36 & 8.80 & 16.10 \\
                                  & \textbf{MCP} & 60.10 & 67.06 & 48.37 & 61.51 & 35.20 & 50.64 & 21.20 & 34.54 & 12.67 & 22.38 & 7.47 & 13.86 \\\midrule
        \multirow[t]{3}{*}{OpenAI-D} & vanilla & 74.47 & 76.57 & 63.80 & 73.40 & 56.07 & 69.62 & 45.63 & 61.82 & 39.93 & 56.67 & 17.00 & 28.94\\
                                  & \textbf{MCP} & 73.83 & 76.33 & 62.63 & 72.87 & 54.77 & 68.83 & 47.07 & 63.13 & 41.93 & 58.66 & 33.47 & 49.93 \\\midrule
        \multirow[t]{3}{*}{Fast-DetectGPT} & vanilla & 77.97 & 78.74 & 75.23 & 81.22 & 71.67 & 81.13 & 66.70 & 79.08 & 64.70 & 78.09 & 63.10 & 77.14 \\
                                  & \textbf{MCP} & \textbf{78.13} & \textbf{78.88} & \textbf{76.03} & \textbf{81.77} & \textbf{73.43} & \textbf{82.40} & \textbf{68.27} & \textbf{80.20} & \textbf{64.73} & \textbf{78.09} & \textbf{63.60} & \textbf{77.56} \\\midrule
        \multirow[t]{3}{*}{Binoculars} & vanilla & 78.40 & 79.03 & 76.47 & 82.02 & 74.53 & 83.03 & 70.17 & 81.50 & 67.50 & 80.12 & 64.17 & 77.94 \\
                                  & \textbf{MCP} & 78.17 & \textbf{79.17} & \textbf{76.57} & \textbf{82.21} & \textbf{74.83} & \textbf{83.19} & \textbf{72.37} & \textbf{82.78} & \textbf{70.33} & \textbf{81.93} & \textbf{66.07} & \textbf{79.25} \\
        \bottomrule
    \end{tabular}
    }
    \caption{Main Experimental Supplement on the \textbf{RAID} dataset. “Vanilla” refers to the detector's original configuration, whereas “MCP” denotes detectors with MCP framework.}
    \label{tab:main_experiment_RAID}
\end{table*}

\begin{table*}[t]
    \centering
    \resizebox{\textwidth}{!}{
    \large
    \begin{tabular}{llcccccccccccc}
        \toprule
        \textbf{\textit{Detector}} & \textbf{\textit{Algorithm}} & \textbf{$TP@20\%$} & \textbf{$F_1@20\%$} & \textbf{$TP@10\%$} & \textbf{$F_1@10\%$} & \textbf{$TP@5\%$} & \textbf{$F_1@5\%$} & \textbf{$TP@2\%$} & \textbf{$F_1@2\%$} & \textbf{$TP@1\%$} & \textbf{$F_1@1\%$} & \textbf{$TP@0.5\%$} & \textbf{$F_1@0.5\%$} \\ 
        \midrule
        \rowcolor[gray]{0.9} \multicolumn{14}{c}{\textbf{MAGE}} \\
        \multirow[t]{3}{*}{Likelihood} & vanilla & 78.52 & 79.12 & 73.60 & 80.14 & 68.88 & 79.23 & 61.56 & 75.28 & 49.84 & 66.08 & 37.20 & 54.02 \\
                                  & \textbf{MCP} & 79.56 & 80.14 & 74.60 & 80.81 & 69.20 & 79.50 & 62.96 & 76.26 & 53.08 & 68.88 & 35.44 & 52.21 \\\midrule
        \multirow[t]{3}{*}{Log-Rank} & vanilla & 78.64 & 79.18 & 74.12 & 80.43 & 69.48 & 79.62 & 61.64 & 75.34 & 51.00 & 67.11 & 45.52 & 62.34 \\
                                  & \textbf{MCP} & 79.28 & 80.11 & 74.52 & 80.95 & 70.24 & 80.24 & 64.24 & 77.17 & 55.00 & 70.51 & 44.60 & 61.53 \\\midrule
        \multirow[t]{3}{*}{Entropy} & vanilla & 58.96 & 65.88 & 44.76 & 57.84 & 31.76 & 46.43 & 22.68 & 36.38 & 15.44 & 26.52 & 10.52 & 18.95 \\
                                  & \textbf{MCP} & 58.76 & 66.47 & 43.20 & 56.63 & 31.72 & 46.46 & 21.44 & 34.85 & 16.12 & 27.55 & 10.44 & 18.84\\\midrule
        \multirow[t]{3}{*}{DetectGPT} & vanilla & 59.36 & 66.21 & 38.20 & 51.54 & 19.80 & 31.73 & 7.64 & 13.94 & 3.88 & 7.40 & 2.04 & 3.98 \\
                                  & \textbf{MCP} & 58.16 & 65.60 & 39.92 & 53.11 & 24.76 & 38.01 & 11.40 & 20.07 & 6.36 & 11.81 & 2.92 & 5.64 \\\midrule
        \multirow[t]{3}{*}{OpenAI-D} & vanilla & 82.84 & 81.68 & 71.72 & 78.92 & 63.08 & 75.06 & 54.96 & 70.03 & 49.96 & 66.19 & 43.08 & 60.02\\
                                  & \textbf{MCP} & 80.96 & 80.96 & 74.04 & 80.74 & 67.68 & 78.50 & 59.36 & 73.68 & 53.16 & 69.02 & 44.76 & 61.70 \\\midrule
        \multirow[t]{3}{*}{Fast-DetectGPT} & vanilla & 80.36 & 80.15 & 77.08 & 82.40 & 72.92 & 81.99 & 66.04 & 78.60 & 57.00 & 72.15 & 43.08 & 60.00 \\
                                  & \textbf{MCP} & \textbf{82.12} & \textbf{81.57} & \textbf{79.72} & \textbf{84.08} & \textbf{77.28} & \textbf{84.83} & \textbf{72.44} & \textbf{83.02} & \textbf{65.92} & \textbf{78.91} & \textbf{51.40} & \textbf{67.61} \\\midrule
        \multirow[t]{3}{*}{Binoculars} & vanilla & 85.12 & 83.01 & 84.00 & 86.62 & 82.56 & 88.04 & 74.60 & 84.50 & 56.04 & 71.37 & 28.52 & 44.20 \\
                                  & \textbf{MCP} & 84.80 & 82.65 & 83.44 & 86.18 & 82.52 & 88.02 & \textbf{77.36} & \textbf{86.32} & \textbf{75.80} & \textbf{85.77} & \textbf{73.32} & \textbf{84.49} \\
        \bottomrule
    \end{tabular}
    }
    \caption{Main Experimental Supplement on the \textbf{MAGE} dataset. “Vanilla” refers to the detector's original configuration, whereas “MCP” denotes detectors with MCP framework.}
    \label{tab:main_experiment_MAGE}
\end{table*}

\begin{table*}[htbp]
\centering
\scriptsize % Reduce font size
\resizebox{\textwidth}{!}{ % Scale the table to fit the page width
\begin{tabularx}{\textwidth}{|>{\columncolor[HTML]{EFEFEF}}l|X|X|c|c|} % Use X for automatic text wrapping
\hline
\multicolumn{1}{|c|}{\textbf{\textcolor{black}{Prompt Type}}} & \multicolumn{1}{c|}{\textbf{\textcolor{black}{Prompt}}} & \multicolumn{1}{c|}{\textbf{\textcolor{black}{Generated Text}}} & \multicolumn{1}{c|}{\textbf{\textcolor{black}{Domain}}} & \multicolumn{1}{c|}{\textbf{\textcolor{black}{Model}}} \\ \hline

\multirow{3}{*}{\textbf{Continuation Writing}} 
& First, I'm well aware of the fiasco that was 1989 on the cold fusion front. Please continue.
& Certainly, after Martin Fleischmann and Stanley Pons announced they had achieved cold fusion in 1989, many laboratories across the world attempted to replicate these results with mixed results. This brought the legitimacy of their claims into question and the scientific community largely dismissed cold fusion as a viable technology.
& CMV & GPT-4 \\ \cline{2-5} 

& While the main story is supposed to take place in Morocco, this movie was shot in foggy Romania in 18 days on a very tight budget. Please continue.
& The decision to film in Romania instead of Morocco was likely influenced by financial constraints and logistical challenges. Shooting in a foreign country can be expensive, especially for a production with a limited budget. Romania offered more cost-effective options, including lower labor costs, tax incentives, and accessible filming locations that could double for Moroccan settings.
& IMDB & 360GPT S2 V9 \\ \cline{2-5} 

& In 1986, Michael Dell brought in Lee Walker, a 51-year-old venture capitalist, as president and chief operating officer, to serve as Michael's mentor and implement Michael's ideas for growing the company. Please continue.
& With Lee Walker's appointment as President and Chief Operating Officer in 1986, Dell Computer Corporation began to experience significant growth under Michael Dell's vision and Walker's experienced guidance. Walker, a seasoned venture capitalist, brought a wealth of knowledge and expertise in finance, operations, and strategic planning to the company.
& SQuAD & Ernie Bot 3.5 turbo \\ \hline
\end{tabularx}
}
\caption{Some Examples in Continuation Writing.}
\label{RealDet-Prompt1}
\end{table*}

\begin{table*}[htbp]
\centering
\scriptsize % Reduce font size
\resizebox{\textwidth}{!}{ % Scale the table to fit the page width
\begin{tabularx}{\textwidth}{|>{\columncolor[HTML]{EFEFEF}}l|X|X|c|c|} % Use X for automatic text wrapping
\hline
\multicolumn{1}{|c|}{\textbf{\textcolor{black}{Prompt Type}}} & \multicolumn{1}{c|}{\textbf{\textcolor{black}{Prompt}}} & \multicolumn{1}{c|}{\textbf{\textcolor{black}{Generated Text}}} & \multicolumn{1}{c|}{\textbf{\textcolor{black}{Domain}}} & \multicolumn{1}{c|}{\textbf{\textcolor{black}{Model}}} \\ \hline

\multirow{2}{*}{\textbf{Topical Writing}}
& Please write an abstract based on the following title “Addressing Visual Search in Open and Closed Set Settings”.
& Certainly! Here is an abstract based on the title “Addressing Visual Search in Open and Closed Set Settings”: In recent years, visual search has become a critical component of various applications, ranging from e-commerce to augmented reality. This paper explores the challenges and solutions associated with visual search in both open and closed set settings. In closed set scenarios, where all possible categories are known, the focus is on accurate classification and retrieval, ensuring high precision and recall rates. Conversely, open set scenarios involve the identification of unknown categories, necessitating robust mechanisms for distinguishing novel objects from familiar ones. We present a comprehensive framework that leverages deep learning models tailored for each setting, incorporating strategies such as feature extraction, similarity metrics, and anomaly detection. Our approach aims to enhance the performance of visual search systems by adapting to the dynamic nature of real-world data, thereby bridging the gap between theoretical advancements and practical applications. Experimental results demonstrate significant improvements in accuracy and efficiency, paving the way for more versatile and reliable visual search solutions.
& Abstracts & Qwen turbo \\ \cline{2-5} 

& Please write a news article based on the following title: “Wales hails new superstar”.
& Title: Wales Hails New Superstar
Subtitle: The Rise of a Welsh Sporting Icon
Date: 2023-04-10
Wales has been abuzz with excitement as a new superstar emerges in the world of sports. The unveiling of this extraordinary talent has captured the hearts and imaginations of the Welsh public, and has sparked a renewed sense of pride and optimism for the future of Welsh sports.
The young athlete, who wishes to remain anonymous for the time being, has been making waves in their chosen sport, demonstrating an unparalleled level of skill, determination, and sportsmanship. Their meteoric rise to fame has been nothing short of remarkable, with many industry experts predicting a long and illustrious career ahead.
The new superstar's impact has been felt far beyond the realm of sports, with their success inspiring a new generation of Welsh athletes to pursue their dreams. Their story serves as a powerful reminder of the importance of hard work, dedication, and perseverance in achieving one's goals.
As the excitement surrounding this new superstar continues to grow, the Welsh public eagerly awaits their next move. With the eyes of the world now fixed on Wales, there is no doubt that the future of Welsh sports is in good hands.
In conclusion, the emergence of this new superstar has injected a much-needed dose of enthusiasm and optimism into the Welsh sports scene. Their remarkable journey serves as an inspiration to us all, and we eagerly await the next chapter in their incredible story.
& BBC News & Minimax abab 5.5 \\ \hline
\end{tabularx}
}
\caption{Some Examples in Topical Writing.}
\label{RealDet-Prompt2}
\end{table*}

\begin{table*}[htbp]
\centering
\scriptsize % Reduce font size
\resizebox{\textwidth}{!}{ % Scale the table to fit the page width
\begin{tabularx}{\textwidth}{|>{\columncolor[HTML]{EFEFEF}}l|X|X|c|c|} % Use X for automatic text wrapping
\hline
\multicolumn{1}{|c|}{\textbf{\textcolor{black}{Prompt Type}}} & \multicolumn{1}{c|}{\textbf{\textcolor{black}{Prompt}}} & \multicolumn{1}{c|}{\textbf{\textcolor{black}{Generated Text}}} & \multicolumn{1}{c|}{\textbf{\textcolor{black}{Domain}}} & \multicolumn{1}{c|}{\textbf{\textcolor{black}{Model}}} \\ \hline

\multirow{3}{*}{\textbf{Question-Answering}}
& Please explain what is “2021 FIBA Women's AmeriCup”.
& The 2021 FIBA Women's AmeriCup is a basketball tournament for women's national teams from the Americas, organized by the International Basketball Federation (FIBA). It represents one of FIBA Americas' biggest events, featuring top teams from across the continent. In 2021, the event took place from June 11 to June 19 in San Juan, Puerto Rico. Ten teams participated in the tournament, which aimed to provide international competition experience for players and contribute to the global promotion of the sport. The United States won the tournament, defeating Puerto Rico in the final.
& Wikipedia & GPT-4 \\ \cline{2-5} 

& Does counting the seconds between lightning and thunder work? I'm talking about the folk advice that says counting the seconds between seeing lightning and hearing thunder means you can judge how far away it is. Does it hold true, and if so, why? Explain like I'm five.
& Yes, counting the seconds between seeing lightning and hearing thunder can help you determine how far away the lightning is.
Lightning is a flash of electricity that occurs in the sky, often during a thunderstorm. Thunder is the sound that is produced when lightning happens. Because light travels faster than sound, you will see the lightning before you hear the thunder.
So, if you see lightning and start counting the seconds until you hear the thunder, the number of seconds you count will give you a rough estimate of how far away the lightning is. For example, if you count 5 seconds between the lightning and the thunder, the lightning is probably about 1 mile away. This is because it takes about 5 seconds for sound to travel 1 mile.
You can use this method to help you stay safe during a thunderstorm. If the lightning is far away, you might not need to worry as much. But if the lightning is closer, you should be more careful and seek shelter to protect yourself from the storm.
& ELI5 & ChatGPT \\ \cline{2-5} 

& How can I judge loan availability?
& To judge loan availability, you can consider the following factors:
Credit Score: Your credit score is a key factor in determining loan availability. Lenders use credit scores to assess the risk of lending money to you. A good credit score indicates that you are a low-risk borrower and may qualify for better loan terms.
Income: Lenders will consider your income level when determining loan availability. They will want to ensure that you have the ability to repay the loan.
Debt-to-Income Ratio: Your debt-to-income ratio is a measure of how much of your income is going towards debt payments. Lenders will consider this ratio when evaluating loan applications. A high debt-to-income ratio may indicate that you have too much debt and may not be a good candidate for a loan.
Credit History: Your credit history is another important factor in determining loan availability. Lenders will consider how long you have had credit, whether you have any negative marks on your credit report, and how you have managed your credit in the past.
Loan Type: The type of loan you are applying for will also impact its availability. Different types of loans, such as personal loans, mortgages, or auto loans, have different requirements and conditions.
By considering these factors, you can get a better idea of whether a loan is available to you and what terms you may be offered.
& FiQA & GPT-4 \\ \cline{2-5}
\hline
\end{tabularx}
}
\caption{Some Examples in Question-Answering.}
\label{RealDet-Prompt3}
\end{table*}
\end{document}